\documentclass[sigconf]{acmart}
\usepackage{amsthm}

\usepackage{amsmath,amssymb}

\usepackage{mathrsfs}
\usepackage{soul}
\usepackage{algorithm}  
\usepackage{algpseudocode}  
\usepackage{balance} 

\AtBeginDocument{%
  \providecommand\BibTeX{{%
    \normalfont B\kern-0.5em{\scshape i\kern-0.25em b}\kern-0.8em\TeX}}}

\AtBeginDocument{%
  \providecommand\BibTeX{{%
    \normalfont B\kern-0.5em{\scshape i\kern-0.25em b}\kern-0.8em\TeX}}}

\setcopyright{acmcopyright}

\copyrightyear{2023} 
\acmYear{2023} 
\setcopyright{acmlicensed}\acmConference[MM '23]{Proceedings of the 31st ACM International Conference on Multimedia}{October 29-November 3, 2023}{Ottawa, ON, Canada}
\acmBooktitle{Proceedings of the 31st ACM International Conference on Multimedia (MM '23), October 29-November 3, 2023, Ottawa, ON, Canada}
\acmPrice{15.00}
\acmDOI{10.1145/3581783.3612006}
\acmISBN{979-8-4007-0108-5/23/10}

\acmSubmissionID{1377}

\begin{document}

\title{Dual-Modal Attention-Enhanced Text-Video Retrieval with Triplet Partial Margin Contrastive Learning}

%

\author{Chen Jiang}
\authornote{Both authors contributed equally to this research.}
\email{qichen.jc@antgroup.com}
\affiliation{%
\institution{Artificial Intelligence Innovation and Incubation Institute, Fudan University}
  \institution{Ant Group}
  \country{} 
}

\author{Hong Liu}
\authornotemark[1]
\email{yizhou.lh@antgroup.com}

\author{Xuzheng Yu}
\email{yuxuzheng.yxz@antgroup.com}
\author{Qing Wang}
\email{wq176625@antgroup.com}
\affiliation{%
  \institution{Ant Group}
  \country{} 
}

\author{Yuan Cheng}
\authornote{Corresponding author.}
\email{cheng_yuan@fudan.edu.cn}
\affiliation{%
  \institution{Artificial Intelligence Innovation and Incubation Institute, Fudan University}
  \country{} 
}

\author{Jia Xu}
\email{steve.xuj@antgroup.com}
\author{Zhongyi Liu}
\email{zhongyi.lzy@antgroup.com}
\affiliation{%
  \institution{Ant Group}
  \country{} 
}

\author{Qingpei Guo}
\authornotemark[2]
\email{qingpei.gqp@antgroup.com}
\author{Wei Chu}
\email{weichu.cw@antgroup.com}

\author{Ming Yang}
\email{m.yang@antgroup.com}
\affiliation{%
  \institution{Ant Group}
  \country{} 
}

\author{Yuan Qi}
\email{qiyuan@fudan.edu.cn}
\affiliation{%
  \institution{Artificial Intelligence Innovation and Incubation Institute, Fudan University}
  \country{} 
}

\renewcommand{\shortauthors}{Chen Jiang et al.}
\begin{abstract}
In recent years, the explosion of web videos makes text-video retrieval increasingly essential and popular for video filtering, recommendation, and search.
Text-video retrieval aims to rank relevant text/video higher than irrelevant ones. The core of this task is to precisely measure the cross-modal similarity between texts and videos.
Recently, contrastive learning methods have shown promising results for text-video retrieval, most of which focus on the construction of positive and negative pairs to learn text and video representations. 
Nevertheless, they do not pay enough attention to hard negative pairs and lack the ability to model different levels of semantic similarity.
To address these two issues, this paper improves contrastive learning using two novel techniques. 
First, to exploit hard examples for robust discriminative power, we propose a novel \textit{Dual-Modal Attention-Enhanced Module (DMAE)} to mine hard negative pairs from textual and visual clues. 
By further introducing a \textit{Negative-aware InfoNCE (NegNCE)} loss, we are able to adaptively identify all these hard negatives and explicitly highlight their impacts in the training loss.
Second, our work argues that triplet samples can better model fine-grained semantic similarity compared to pairwise samples. 
We thereby present a new \textit{Triplet Partial Margin Contrastive Learning (TPM-CL)} module to construct partial order triplet samples by automatically generating fine-grained hard negatives for matched text-video pairs.
The proposed TPM-CL designs an adaptive token masking strategy with cross-modal interaction
to model subtle semantic differences.
Extensive experiments demonstrate that the proposed approach outperforms existing methods on four widely-used text-video retrieval datasets, including MSR-VTT, MSVD, DiDeMo and ActivityNet. Code is publicly available at \href{https://github.com/alipay/Ant-Multi-Modal-Framework}{https://github.com/alipay/Ant-Multi-Modal-Framework}.
\end{abstract}

\ccsdesc[500]{Information systems~Novelty in information retrieval}

\keywords{Text-Video Retrieval, Dual-Modal Attention-Enhanced, Negative-aware InfoNCE, Triplet Partial Margin Contrastive Learning}



\maketitle

\section{Introduction}
With the explosive growth of videos in recent years, the task of text-video retrieval has become increasingly essential and popular.
The goal of text-video retrieval is to retrieve videos that are most semantically relevant to the given text query.
A typical paradigm tends to first embed texts and videos into a joint latent space and then employ a distance metric to measure cross-modal similarity~\cite{HERO,univl2020,MMT,clip4clip2021,clip2video2021}. 
A critical challenge is to learn precise semantic similarities between texts and videos. 
The recent trend towards large-scale contrastive image-language pre-training like CLIP~\cite{clip2021} mitigates this issue to some extent~\cite{clip4clip2021,ts2net2022,xclip2022},
yet they tend to neglect the distinct role of hard examples, 
leading to confusion with hard positives/negatives and noisy correspondence. 
Moreover, most existing contrastive learning works focus on the pairwise semantic relation, which lacks the ability to measure different levels of semantic similarity~\cite{zhang2022ArcCSE}.


\begin{figure}[ht!] 
  \centering
  \includegraphics[width=1.0\linewidth]{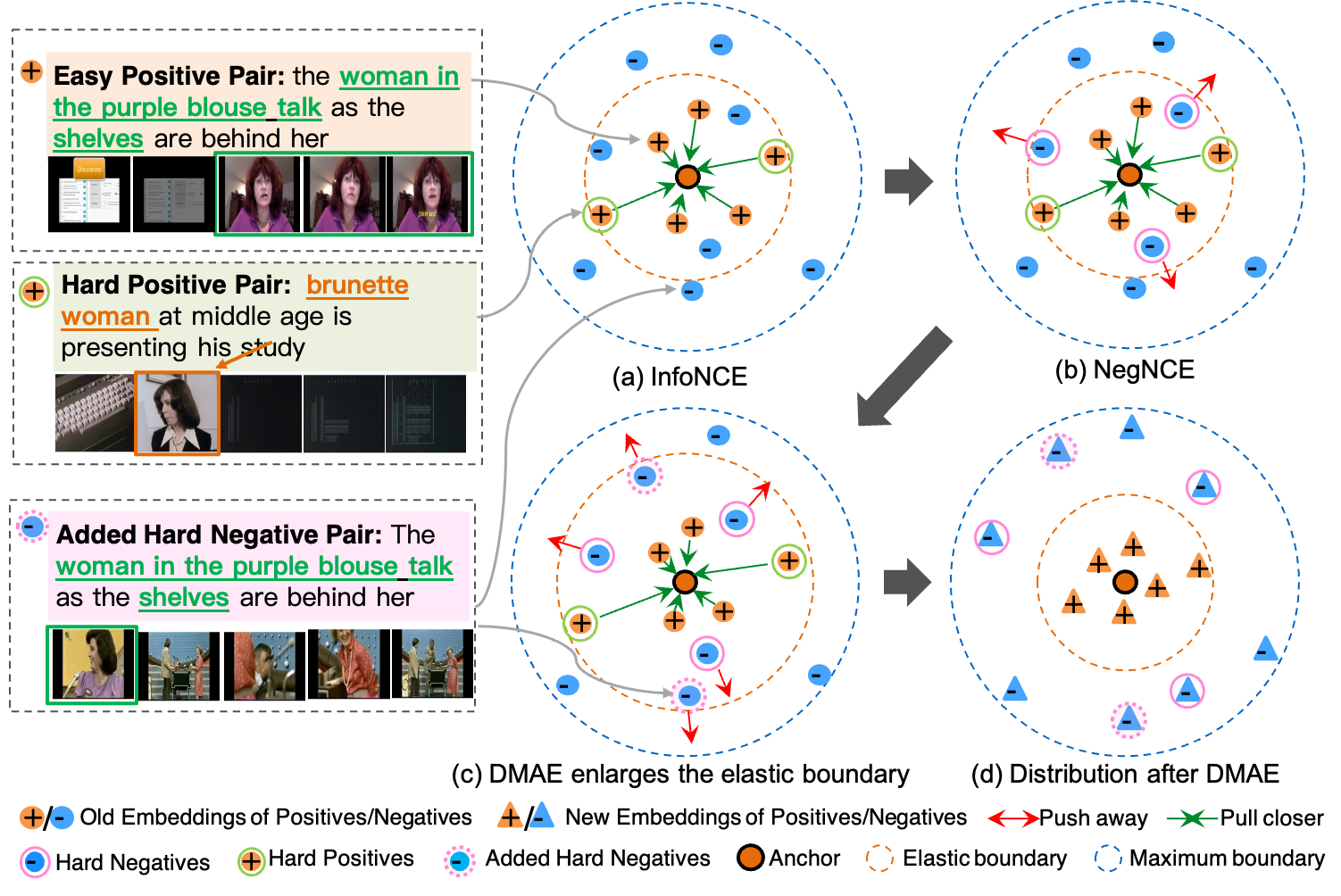}
  \caption{Illustration of embedding distributions constrained by NegNCE and DMAE. 
  (a) Embedding distribution constrained by InfoNCE loss. Hard negative pairs that lie inside the elastic boundary are ignored. (b) Hard negative pairs inside the elastic boundary are adaptively incorporated into the NegNCE loss, and are pushed away from the anchor. (c) DMAE enlarges the elastic boundary and adds more hard negatives. (d) After training with DMAE, positive and negative pairs are further pushed away from each other. }
  \label{fig:dmae}
\end{figure}

Contrastive learning becomes a popular representation learning paradigm for text-video retrieval recently~\cite{taco2021,univl2020,frozen2021,Li2021AlignAP,clip4clip2021,videoclip2022,xpool2022,xclip2022,ts2net2022}. 
Among them, the majority~\cite{videobert2019,univl2020,MMT,hanet2021,clip4clip2021,xpool2022,xclip2022,ts2net2022} rely on conventional pairwise contrastive losses (\textit{e.g.}, NCE~\cite{nce2010}, BCE~\cite{hanet2021}, infoNCE~\cite{infonce2018}) to learn a cross-modal embedding space, which minimizes the distance of matched pairs while maximizing the distance of all other negative pairs in a batch.
As shown in Fig.~\ref{fig:dmae}(a), the elastic boundary is defined by the farthest positives, and hard negatives refer to negatives that lie inside the elastic boundary and are closer to the anchor than the farthest positives. 
Usually, these conventional contrastive losses focus on positives and treat all negatives in the batch equally, 
without distinguishing between hard and easy ones.
However, hard negatives should make a greater impact on the discrimination between matched and mismatched pairs. 
Most of the current approaches adopt random sampling strategies~\cite{actbert2020,univl2020} or a specific sampling strategy~\cite{taco2021,hanet2021,acp2022} to cut the number of negatives to a fixed number. 
These strategies may result in sub-optimal learning or overlooking some hard negatives. This is because the limited number of selected negatives may not accurately reflect the true distribution of negative pairs.
To address these issues, we introduce a \textit{Negative-aware InfoNCE (NegNCE)} loss to adaptively find out all hard negative pairs inside the elastic boundary and incorporate them into the training objective as in Fig.~\ref{fig:dmae}(b). 

Nevertheless, the fundamental challenge is: "How to select as many hard negatives as possible?". Selecting hard negatives is not as straightforward as selecting positives.
As in Fig.~\ref{fig:dmae}(b), it is easy to miss the hard negative pairs near the elastic boundary as they are more challenging to differentiate from the positives.
Previous work~\cite{tupleinfonce2021} has observed that \textit{"strong variations between the positive and anchor samples usually result in smaller shared information but a greater degree of invariance against nuisance variables"}.
Therefore, 
we need to enhance text-video pairs so that contrastive learning can keep the shared information between positives and anchors while mining hard negatives.
Here, we propose a novel \textit{Dual-Modal Attention-Enhanced Module (DMAE)} to enlarge variations between easy and hard positives so that similar hard negatives that lie near the elastic boundary can be extracted  as in Fig.~\ref{fig:dmae}(c).
As the cases shown in Fig.~\ref{fig:dmae}, 
when matching visual content to a text query, we categorize text-video pairs accordingly. Those pairs with multiple frames matching the query are considered as \textit{easy positives}, while those with only a single frame match are classified as \textit{hard positives}.
By enlarging the discrimination between easy positives and hard positives, we can find those \textit{hard negatives} with single-frame visual content that only partly matches the text query.
Specifically, DMAE enhances text-video pairs through two components named \textit{Textual Attention} and \textit{Visual Attention} to find out more challenging hard negatives while filtering out easy negatives. 
In this way, we expect that positives are pulled closer to each other while negatives are pushed away after training as in Fig.~\ref{fig:dmae}(d).

As mentioned above, most existing works~\cite{clip4clip2021,ts2net2022,xclip2022,xpool2022,drl2022} focus on pairwise contrastive losses. Yet using pairwise losses essentially applies a binary quantization on the semantic similarity among text-video pairs, \emph{i.e.}, to either positive or negative pairs, which is a very coarse way to measure their relations. In contrast, we prefer to have a finer measurement on the semantic similarity among text-video pairs so as to take advantage of different levels of semantic similarity in contrastive learning. As the case in the right part of Fig.~\ref{fig:framework}, the original text query with more details of the video should be more similar than the masked one \textit{"the woman talk as the shelves are behind her"} or \textit{"the woman talk"} to the video. 
Therefore, we propose the \textit{Triplet Partial Margin Contrastive Learning (TPM-CL)} module to model the subtle difference in semantic similarity by leveraging partial order triplet samples. {Unlike previous work~\cite{falcon_relevance-based_2022} which adopts a relevance-based margin in the triplet loss to impart subtle semantic differences to the model, our focus is on the automatic generation of partial order triplet samples.}
Previous works construct triplet samples by offline text token masking for text matching~\cite{zhang2022ArcCSE} or in-batch hard negative mining for face recognition~\cite{facenet2015,triplet4reid2020}. Instead, we design an automatic scheme to generate partial text-video triplets by cross-modal interaction. 
Then an auxiliary target based on triplet ranking loss is adopted to consume the fine-grained semantic similarity among the triplet samples.

Extensive experiments on four text-video retrieval benchmarks show that the proposed method achieves the state-of-the-art performance, including MSR-VTT (212.2 rsum), MSVD (209.3 rsum), DiDeMo (206.3 rsum) and ActivityNet (207.5 rsum). Our approach outperforms the previous SOTA methods by +2.2\%, +0.7\%, +2.4\%, +2.7\% absolute improvements on these benchmarks. The ablation experiments demonstrate that the proposed DMAE and TPM-CL modules both improve the text-video retrieval performance. 

Our main contributions can be summarized as follows:
\begin{itemize}
    \item We propose a novel \textit{Dual-Modal Attention-Enhanced Module (DMAE)} to leverage hard negatives from textual and visual clues, and introduce a \textit{Negative-aware InfoNCE (NegNCE)} loss to explicitly incorporate these hard negatives into the training objective.
    \item We present a new \textit{Triplet Partial Margin Contrastive Learning (TPM-CL)} module to automatically generate partial order triplet samples by an adaptive token masking strategy with cross-modal interaction and model different levels of semantic similarity among them.
    \item We report top performance of retrieval performance on four text-video retrieval benchmarks and conduct extensive ablation studies to demonstrate the merits of our approach.
\end{itemize}

\section{Related Work}
\subsection{Text-Video Retrieval}
Most of the existing works directly apply the pre-trained backbone to obtain textual and visual representations, followed by interaction modules to measure the cross-modal similarity. 
With the success in many downstream tasks~\cite{COOP,CLIPCAP,STYLECLIP,DALL_E2,ActionCLIP}, CLIP~\cite{clip2021} has injected new impetus into the improvement of text-video retrieval and quickly becomes one of the mainstream backbones~\cite{clip4clip2021,xpool2022,clip2tv2021,camoe2021,clip2video2021,drl2022,xclip2022}.
For example, CLIP4CLIP~\cite{clip4clip2021} and CLIP2TV~\cite{clip2tv2021} transfer image knowledge to text-video retrieval to learn better representations. 
TS2-Net~\cite{ts2net2022} and CenterCLIP~\cite{centerclip2022} introduce a token selection or token clustering module to find the most informative tokens. 
XCLIP~\cite{xclip2022} first applies multi-grained contrastive learning to reduce the negative effects of unnecessary information. DRL~\cite{drl2022} proposes an effective interaction method to solve the sequential matching problem, and an auxiliary loss to reduce feature redundancy.
Yet these works do not pay enough attention to either the entailment relation among hard examples or triplet samples.
Our work also applies contrastive learning under the aforementioned typical paradigm. 
Differently, we are the first to improve the discriminative power from pairwise and triple-wise perspectives by hard negative mining and automatic partial order triplet generating.



\subsection{Negative Mining in Contrastive Learning}

Most of the negative mining methods can be divided into two categories: negative sampling and negative generating. The former focuses more on selecting hard negatives from a given corpus, while the latter aims to generate hard negatives in certain ways.

In terms of negative sampling, the majority of current methods~\cite{actbert2020,univl2020,frozen2021,clip4clip2021,videoclip2022,xclip2022,ts2net2022} use random sampling strategies.
TACo~\cite{taco2021} utilizes a token-aware cascade hard negative sampling strategy to select a fixed number of hard negatives within a batch. 
Moreover, triplet ranking loss with online triplet mining often acts as an auxiliary target to guide the text-video alignment, which usually selects the hardest negative sample to construct triplet samples~\cite{Dong2021DualEF,acp2022,hanet2021}.
Nevertheless, these strategies may result in sub-optimal learning or missing some hard negatives
as the distribution of hard pairs may be either scarce or dense, depending on the batch size.
Some other works~\cite{hit2021,lgdn2022, sstvlmss2022} introduce a momentum mechanism (like MOCO~\cite{moco2019}) to maintain a large negative queue as the corpus of negatives. Although MOCO-based methods can decouple the number of negative samples from the batch size, they require keeping a large memory up-to-date for negatives.
Different from them, we focus on adaptively finding out hard negative pairs according to the distribution of pairs and taking them into consideration in the pairwise contrastive loss.

As for negative generating, ~\cite{Wang2022LearnTU} proposes an offline strategy to generate negated text-video pairs by partially negating its original caption, which is unable to model the negation from visual clues. 
Authors in~\cite{MoCHi2020} adopt a feature mix-up strategy to generate hard negatives, which may lead to false negatives. Yet the major drawback of these methods is the lack of cross-modal interaction.  
This work generates fine-grained hard negatives by an adaptive token masking strategy with cross-modal interaction to construct triplet samples.
Coupling with the triplet ranking loss, it is able to model different levels of semantic similarity among them.

\section{Method}
\begin{figure}[ht!] 
  \centering
  \includegraphics[width=0.9\linewidth]{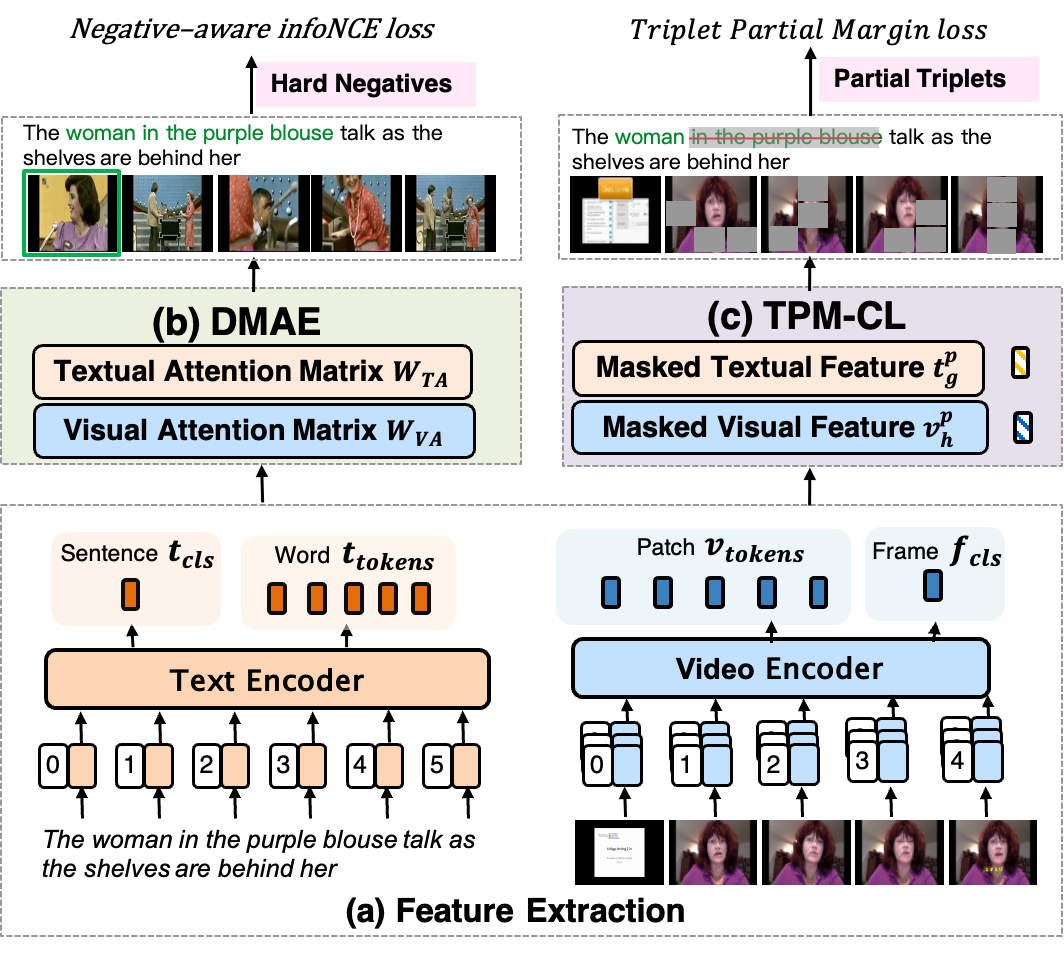}
  \caption{Overview of our approach, containing two major modules: (1)\textit{Dual-Modal Attention-Enhanced Module
(DMAE)}, which aims to mine hard negatives and is coupled with a \textit{Negative-aware InfoNCE(NegNCE)} loss to incorporate these hard negatives into training objective, and (2)\textit{Triplet Partial Margin
Contrastive Learning (TPM-CL)}, which aims to model the partial order of semantics among triplet samples.}
  \label{fig:framework}
\end{figure}

This section presents each component of the proposed method (Fig.~\ref{fig:framework}). Starting with an introduction of feature representation in Sec.~\ref{subsec:feats}, we then elaborate on the details of our two core modules:
(i) \textit{Dual-Modal Attention-Enhanced Module (DMAE)}, (ii) \textit{Triplet Partial Margin Contrastive Learning (TPM-CL)}, in Sec.~\ref{subsec:dmae} and~\ref{subsec:TPM-CL}, respectively, followed by the total objective function in Sec.~\ref{subsec:obj}.

\subsection{Feature Representation}\label{subsec:feats}
Given a text set $\mathcal{T}$ and a video set $\mathcal{V}$, our target is to learn a similarity function $f(t_i, v_i)$, which calculates the similarity score between a text $t_i \in \mathcal{T}$ and a video $v_i \in \mathcal{V}$.
Following the typical text-video retrieval framework~\cite{clip4clip2021, clip2tv2021,clip2video2021}, our model is composed of a text encoder $g$ and a video encoder $h$, which leverages CLIP~\cite{clip2021} as a backbone. The text encoder $g(t_i)$ produces the sentence-level textual feature $\mathbf{t}_{cls} \in \mathbb{R}^{1 \times D}$ and the word-level textual feature $\mathbf{t}_{tokens} \in \mathbb{R}^{M \times D}$, where $M$ is the length of $t_i$ and $D$ is the dimension of features. The video encoder $h(v_i)$ produces the frame-level visual feature $\mathbf{f}_{cls} \in \mathbb{R}^{N \times D}$ and the patch-level visual feature $\mathbf{v}_{tokens} \in \mathbb{R}^{P \times D}$, where $N$ is the number of frames and $P$ is the length of the patch sequence.
Note that $\mathbf{f}_{cls}$ and $\mathbf{v}_{tokens}$ are extracted from separate frames and the interaction among frames is ignored. Thus, we further use a temporal encoder to aggregate the features of all frames as in previous works ~\cite{clip4clip2021,ts2net2022,xclip2022,xpool2022}. Then, we obtain the aggregated frame-level visual feature $\mathbf{v}_{h} \in \mathbb{R}^{N \times D}$.
\subsection{Dual-Modal Attention-Enhanced Module}\label{subsec:dmae}

\begin{figure}[ht!] 
  \centering
\includegraphics[width=0.9\linewidth]{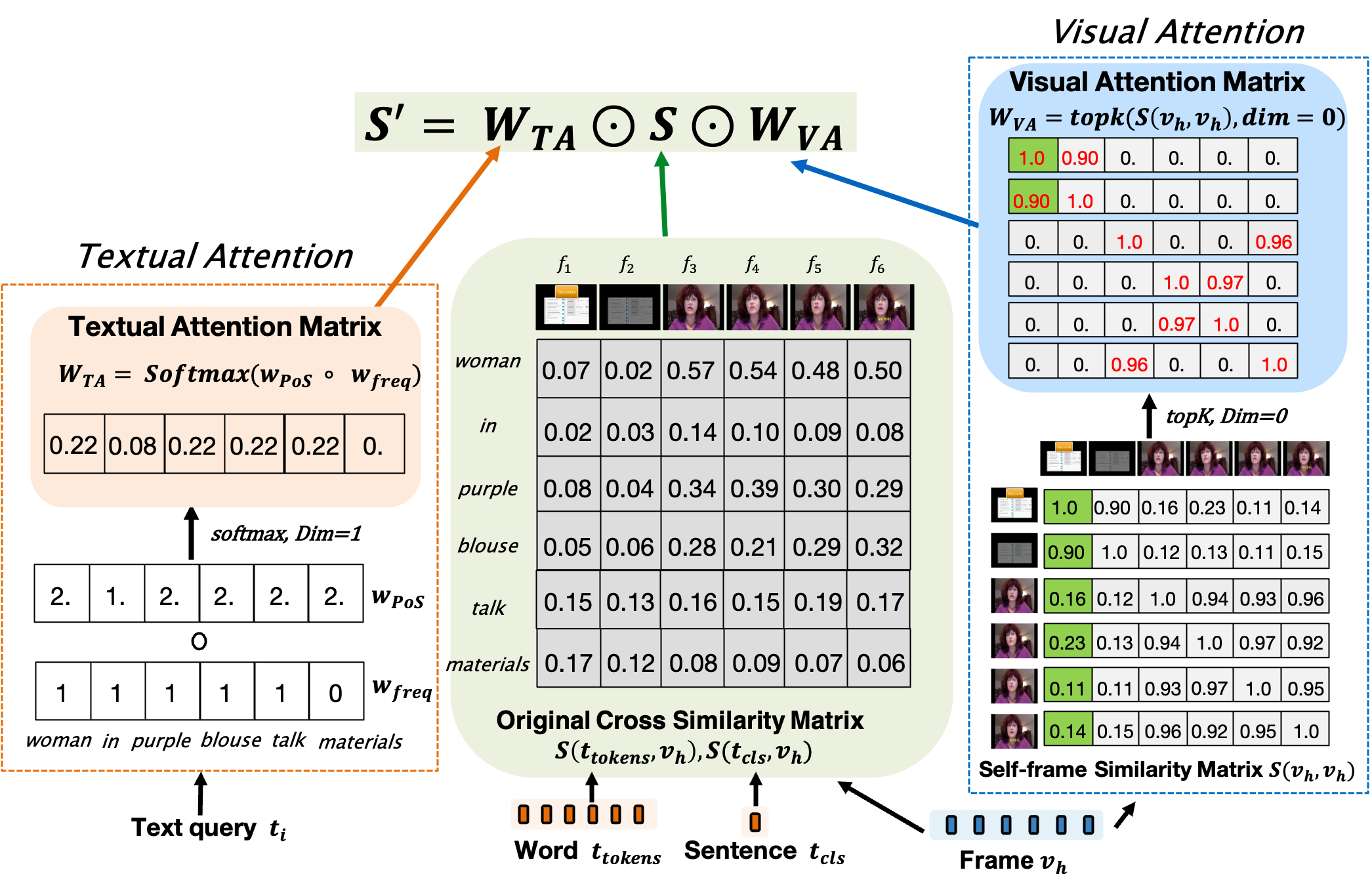}
  \caption{Illustration of DMAE, which mines hard negatives from textual and visual clues}.
  \label{fig:dmae_diagram}
\end{figure}

To alleviate the limitations of InfoNCE loss for overlooking hard negatives, we present a modified \textit{Negative-aware InfoNCE (NegNCE)} loss. 
In addition, we introduce a novel \textit{Dual-modal Attention-Enhanced Module (DMAE)} to optimize representations of text-video pairs, aiming to find out more challenging hard negatives. 
As shown in Fig.~\ref{fig:dmae_diagram}, DMAE consists of two components, which are 1) \textit{Textual Attention}, aiming to mine crucial textual clues; and 2) \textit{Visual Attention}, aiming to explore the intrinsic characteristics from visual clues. 
Then we obtain the \textit{Textual Attention Matrix $W_{TA}$ and Visual Attention Matrix $W_{VA}$}, which are applied to incorporate the crucial textual and visual clues into the final similarity calculation. After that, we get the attention-enhanced similarity matrix $\mathcal{S'}$ and employ the NegNCE loss to train our model.

$\textbf{Textual Attention}$. \label{subsec:dmae_text}
Some works~\cite{finegrained2019,Chen2020FineGrainedVR,taco2021} have observed that content words with specific PoS tags, such as nouns and verbs, are more likely than function words to be aligned with visual content in the video. Moreover, words with a high frequency in a paragraph also tend to show higher relevance to videos. Hence, our idea is to obtain two weight vectors $\mathbf{w}_{Pos}$ and $\mathbf{w}_{freq}$ from these two aspects for modeling the crucial textual clues.
The algorithm is shown in Algorithm~\ref{alg:algo_ta}. Next, as shown in the left part of Fig.~\ref{fig:dmae_diagram}, we construct the \textit{Textual Attention Matrix $W_{TA}$} as follows:
   \begin{equation}
    \mathbf{W}_{TA} = Softmax(\mathbf{w}_{PoS} \circ \mathbf{w}_{freq}) \in \mathbb{R}^{1 \times M},
    \label{eq:wta}
\end{equation}
where $\circ$ denotes element-wise multiplication.

\begin{algorithm}[htb]  
  \caption{Textual Attention.}  
  \label{alg:algo_ta}  
  \begin{algorithmic}[1]  
    \Require  
      A text query $t_i$ with $M$ words of the video $v_i$, $t_i = [s_1,s_2,...,s_M]$;  
      All description sentences of the video $v_i$;
    \Ensure  
      A weight vector of PoS, $\mathbf{w}_{PoS} = [p_{1},p_2,...,p_M]$; 
      A weight vector of word frequency, $\mathbf{w}_{freq} = [q_1,q_2,...,q_M]$; 
    \State Extracting the PoS of each word with the \textit{Spacy}~\cite{spacy2} toolkit; 
    \State Defining the significant PoS set: $PSIG$ = ['NOUN','VERB','ADJ'];
    \State Concatenating all sentences into a paragraph $T$;
    \State Calculating the word frequency with the \textit{TF-IDF} method~\cite{SprckJones2021ASI,Aizawa2003AnIP};
    \State Selecting the irrelevant word set $\mathcal{F}$: the $k$ words with the lowest tf-idf score in $T$;
    \For{$m=1$ to $M$}
        \If{ PoS of word $s_m \in PSIG$}  \State $\mathbf{p_m} \gets \eta (\eta>1)$; \Comment{$\eta=2$ by default}
        \Else \State $\mathbf{p_m} \gets 1$;
        \EndIf
        \If{word $s_m \in \mathcal{F}$} \State $\mathbf{q_m} \gets 0.$;
        \Else \State $\mathbf{q_m} \gets 1.$;
        \EndIf
    \EndFor 
    \label{code:fram:select} \\  
    \Return $\mathbf{w}_{PoS}$, $\mathbf{w}_{freq}$;  
  \end{algorithmic}  
\end{algorithm}

$\textbf{Visual Attention}$. \label{subsec:dmae_vis}
Due to the redundancy nature in continuously changing visual frames, there often exists more than one critical frame. Some recent works \cite{centerclip2022,ts2net2022} apply a token selection algorithm to reduce the redundant visual tokens, which may abandon informative tokens due to the limited number of selected tokens. Differently, our work argues that shared information of critical frames can facilitate representations as well. 
Towards this end, we aim to enhance samples by aggregating the shared information of critical frames.

As shown in the right part of Fig.~\ref{fig:dmae_diagram}, we first utilize cosine similarity to compute the similarities between frames based on $\mathbf{v}_h$. Then, we obtain the self-frame similarity matrix $\mathcal{S}(\mathbf{v}_{h}, \mathbf{v}_h) \in \mathbb{R}^{N \times N}$ to capture the intrinsic similarity relations among frames. 
Next, we build the \textit{Visual Attention Matrix $W_{VA}$} as follows:
\begin{equation}
    W_{VA} = topK(\mathcal{S}(\mathbf{v}_{h}, \mathbf{v}_h), dim=0) \in \mathbb{R}^{N \times N},
\end{equation}
where topK is set to top-2 by default. $W_{VA}$ preserves similarities between the two most similar frames and erases the others to 0.

After obtaining $W_{TA}$ and $W_{VA}$, we construct the attention-enhanced similarity matrix $\mathcal{S'}$ for each text-video pair $(t_i,v_i)$ as follows:
\begin{flalign}
\begin{split}
    & \mathcal{S'}(\mathbf{t}_{tokens}, \mathbf{v}_h) = W_{TA} \odot \mathcal{S}(\mathbf{t}_{tokens}, \mathbf{v}_h) \odot W_{VA}, \\ 
    & \mathcal{S'}(\mathbf{t}_{cls}, \mathbf{v}_h) = \mathcal{S}(\mathbf{t}_{cls}, \mathbf{v}_h) \odot W_{VA}, \\ 
    & \mathcal{S'} = \frac{1}{2}(\mathcal{S'}(\mathbf{t}_{tokens}, \mathbf{v}_h) +  \mathcal{S'}(\mathbf{t}_{cls}, \mathbf{v}_h)) \in \mathbb{R}^{1 \times N}, 
\end{split}
\label{eq:pos_neg2}
\end{flalign}
where  $\odot$ means dot-product, $\mathcal{S}$ is the original cross similarity matrix calculated based on the input textual features and visual features.

Finally, we apply the \textbf{T}oken-wise \textbf{I}nteraction(\textbf{TI}) or the \textbf{W}eighted \textbf{T}oken-wise \textbf{I}nteraction(\textbf{WTI}) method \cite{drl2022} on $\mathcal{S'}$ to get the final similarity score $sim(t_i,v_i)$ of each pair $(t_i,v_i)$.


$\textbf{Negative-aware InfoNCE Loss}$. 
Most prior works~\cite{clip2021,ts2net2022,xclip2022,drl2022} adopt the symmetric InfoNCE loss to optimize the retrieval model, which only considers the positive pairs (${t_i,v_i}$) with little attention to the hard negative pairs (${t_i,v_j}) (i \neq j)$. The InfoNCE loss can be formulated as:  
\begin{equation}
\begin{split}
     \mathcal{L}_p^{t2v}  = - \frac{1}{B}\sum_{i}^{B}log\frac{exp(\tau \cdot sim(t_i, v_i))}{\sum_{j=1}^{B}exp(\tau \cdot sim(t_i,v_j))}, \\
     \mathcal{L}_p^{v2t}  = - \frac{1}{B}\sum_{i}^{B}log\frac{exp(\tau \cdot sim(t_i, v_i))}{\sum_{j=1}^{B}exp(\tau \cdot sim(t_j,v_i))}.
     \label{eq:pos_nce}
\end{split}
\end{equation}

Different from the above InfoNCE loss, we propose a modified \textit{Negative-aware InfoNCE (NegNCE)} loss, {which identifies all hard negatives within a batch and penalizes them more heavily in the training loss.}


In order to adaptively find out the hard negative pairs, we additionally compute a marginal similarity score $sim_{ij}^{m}$ for all pairs $(t_i,v_j)$ in a batch of $B$ pairs, which is expected to measure the distances between the hard negative and positive pairs. Concretely, the marginal similarity score is calculated by: 
\begin{equation}
\begin{split}
    sim_{ij}^{m} = max(0, \ sim(t_i,v_j) - sim(t_i,v_i) + \xi) \\
    + max(0, \ sim(t_j, v_i) - sim(t_i,v_i) + \xi), \forall i,j \in [1,B], \label{eq:neg_mask}
\end{split}
\end{equation}
where $\xi$ is the margin and is set to 0 by default. $sim_{ij}^{m}$ denotes that when the similarity of the negative pair $(t_i,v_j)$ is larger than the similarity of the positive one $(t_i,v_i)$, it equals the difference between them, otherwise it will be set to zero.
Therefore, if $sim_{ij}^{m} >0$, we set the pair $(t_i,v_j) \in \mathcal{N}$, where $\mathcal{N}$ is the set of hard negative pairs and represents all negatives inside the elastic boundary in Fig.\ref{fig:dmae}(c).

Next, the effect of hard negative pairs can be measured as:
\begin{equation}
\begin{split}
     \mathcal{L}_n^{t2v}  = -\frac{1}{H}\sum_{(t_i,v_j)\in \mathcal{N}} log(1-p_{ij}^{t2v}),\\
     \mathcal{L}_n^{v2t}  = -\frac{1}{H}\sum_{(t_i,v_j)\in \mathcal{N}} log(1-p_{ij}^{v2t}),
     \label{eq:neg_nce}
    \end{split}
\end{equation}
where $H$ is the number of negative pairs in $\mathcal{N}$. The symmetric probabilities $p_{ij}^{t2v}$ and $p_{ij}^{v2t}$ of each pair $(t_i,v_j)$ are computed by: 
\begin{equation}
\begin{split}
   p_{ij}^{t2v} = \frac{exp(\tau \cdot sim(t_i, v_j))}{\sum_{k=1}^{B}exp(\tau \cdot sim(t_i,v_k))}, \forall i,j \in [1,B],\\
   p_{ij}^{v2t} = \frac{exp(\tau \cdot sim(t_i, v_j))}{\sum_{k=1}^{B}exp(\tau \cdot sim(t_k,v_j))}, \forall i,j \in [1,B].
\end{split}
\end{equation}


Finally, we compute the symmetric weighted loss based on the corresponding positive and negative pairs as follows:
\begin{equation}
\begin{split}
     \mathcal{L}^{t2v}  = \gamma_1 \cdot \mathcal{L}_p^{t2v} + \gamma_2 \cdot \mathcal{L}_n^{t2v}, \\
     \mathcal{L}^{v2t}  = \gamma_1 \cdot \mathcal{L}_p^{v2t} + \gamma_2 \cdot \mathcal{L}_n^{v2t},
     \label{eq:pos_neg}
\end{split}
\end{equation}

\begin{equation}
     \mathcal{L}_{NegNCE}  = \frac{1}{2}(\mathcal{L}^{t2v}+\mathcal{L}^{v2t}),
     \label{eq:pos_neg2}
\end{equation}
where $\gamma_1$ and $\gamma_2$ are the weighting parameters.

\subsection{Triplet Partial Margin Contrastive Learning}\label{subsec:TPM-CL}
\begin{figure}[ht!] 
  \centering
  \includegraphics[height=50mm,width=0.9\linewidth]{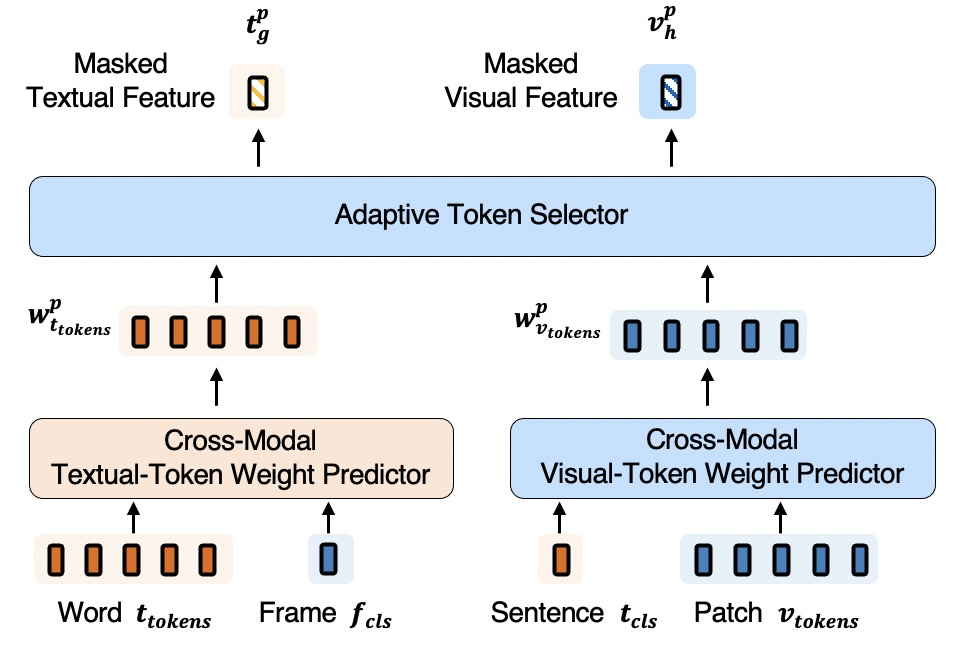}
  \caption{Illustration of \textit{TPM-CL}, which generates partial order triplet samples with cross-modal interaction and models the subtle difference in semantics among them.}
  \label{fig:TPM-CL_diagram}
\end{figure}

In this section, we elaborate on the details of the proposed \textit{Triplet Partial Margin Contrastive Learning (TPM-CL)} module. In order to capture different levels of semantic similarity, the TPM-CL module automatically generates partial order triplet samples for matched text-video pairs and optimizes an auxiliary \textit{Triplet Partial Margin (TPM)} loss. As shown in Fig.~\ref{fig:TPM-CL_diagram}, TPM-CL is formed by two key components, namely, 1)\textit{Cross-Modal Token Weight Predictor} and 2)\textit{Adaptive Token Selector}. The former aims to utilize the cross-modal interaction to predict token weights. The latter generates partial triplets by masking informative textual and visual tokens according to their weights. At last, we design an auxiliary target based on triplet ranking loss to learn the similarity levels.

$\textbf{Cross-Modal Token Weight Predictor}$. Intuitively, the importance of the visual patches and the text words differs depending on the given context. Thus, we use coarse-grained sentence-level and frame-level features to select the most informative tokens. 

Given the word-level textual feature $\mathbf{t}_{tokens}$ and the frame-level visual feature $\mathbf{f}_{cls}$, we first apply a linear projection layer (an MLP) over $\mathbf{f}_{cls}$ for dimension alignment and output $\hat{\mathbf{f}}_{cls} \in \mathbb{R}^{M \times D}$. 
We then concatenate $\mathbf{t}_{tokens}$ with $\hat{\mathbf{f}}_{cls}$. Finally, we feed the concatenated feature to an adaptive module $f_{tw}(\cdot)$ to calculate the weight of each textual token:
\begin{equation}
    \mathbf{w}_{t_{tokens}}^{p} = f_{tw}([\mathbf{t}_{tokens};MLP(\mathbf{f}_{cls})]) \in \mathbb{R}^{1 \times M},
    \label{eq:text_token_weights}
\end{equation}
where $f_{tw}(\cdot)$ is composed of another MLP and a Softmax layer.

In the same manner, we use the sentence-level textual feature $\mathbf{t}_{cls}$ and the patch-level visual feature $\mathbf{v}_{tokens}$ to calculate the element-wise weight of each visual token, which can be formulated as:
\begin{equation}
    \mathbf{w}_{v_{tokens}}^{p} = f_{vw}([\mathbf{v}_{tokens};MLP(\mathbf{t}_{cls})]) \in \mathbb{R}^{ 1 \times P},
    \label{eq:visual_token_weights}
\end{equation}
where $f_{vw}$ has a similar structure to $f_{tw}$.

$\textbf{Adaptive Token Selector}$. After obtaining the weight of each textual and visual token, we adopt an adaptive approach to generate a triplet with entailment relation. We first mask the original textual and visual features according to the binary masks as follows:
\begin{equation}
    \mathbf{t}_{tokens}^{p} = \mathbf{t}_{tokens} \circ \mathbf{b}_t \in \mathbb{R}^{M \times D},
    \label{eq:th_tokens}
\end{equation}
\begin{equation}
    \mathbf{v}_{tokens}^{p} = \mathbf{v}_{tokens} \circ \mathbf{b}_v \in \mathbb{R}^{P \times D},
    \label{eq:vh_tokens}
\end{equation}
where $\circ$ denotes element-wise multiplication with broadcasting, $\mathbf{b}_t = \{b_i^{r_i}\} \in \mathbb{R}^{ 1 \times M} $ and $\mathbf{b}_v = \{b_j^{r_j}\} \in \mathbb{R}^{ 1 \times P}$ are the binary masks for the textual and visual tokens, respectively. The element $b_i^{r_i}$ in $\mathbf{b}_t$ or $\mathbf{b}_v$ is defined as:
\begin{equation}
    b_i^{r_i} = \begin{cases}
    1, & cum\_sum(r_i) < \tau, \\
    0, & cum\_sum(r_i) > \tau,
    \end{cases}
\label{eq:triplet_mask}
\end{equation}
where $cum\_sum(r_i)$ means the cumulative weights till the element $b_i$ which ranks $r_i$th in descending order. And $\tau$ is a fixed threshold that indicates the ratio of features to be masked.

We then derive the weighted global textual feature:
\begin{equation}
    \mathbf{t}_g = \mathbf{t}_{tokens} \odot \mathbf{w}^p_{t_{tokens}} \in \mathbb{R}^{1 \times D},
    \label{eq:tg}
\end{equation}
and the corresponding masked textual feature:
\begin{equation}
    \mathbf{t}_g^p = \mathbf{t}_{tokens}^{p} \odot \mathbf{w}^p_{t_{tokens}} \in \mathbb{R}^{1 \times D},
    \label{eq:tgh}
\end{equation}
where operation $\odot$ in Eq.~\ref{eq:tg}-\ref{eq:tgh} means dot-product. 

Finally, we apply a temporal encoder like in Sec.~\ref{subsec:feats} to aggregate $\mathbf{v}_{tokens}^{p}$ to obtain the masked visual feature $\mathbf{v}_h^p \in \mathbb{R}^{N \times D}$.

$\textbf{Triplet Partial Margin Loss}$. In order to model cross-modal partial order of semantics, we formulate the margin losses as:
\begin{equation}
\begin{split}
    \mathcal{L}_{trpl, 1}= & max(0, -\ \mathcal{S}(\mathbf{t}_{cls}, \mathbf{v}_h) + \mathcal{S}(\mathbf{t}_{cls}, \mathbf{v}_{h}^{p}) + \delta),  \\
    \mathcal{L}_{trpl, 2}= & max(0, -\ \mathcal{S}(\mathbf{t}_{g}, \mathbf{v}_h) + \mathcal{S}(\mathbf{t}_{g}, \mathbf{v}_{h}^{p}) + \delta),  \\
    \mathcal{L}_{trpl, 3}= & max(0, -\ \mathcal{S}(\mathbf{t}_{g}, \mathbf{v}_h) + \mathcal{S}(\mathbf{t}_{g}^{p}, \mathbf{v}_h) + \delta),   \\
\end{split}
\label{eq:triplet_loss}
\end{equation}
where $\delta$ is the margin constant. 

Finally, the auxiliary \textit{Triplet Partial Margin (TPM)} loss is formulated as:
\begin{equation}
    \mathcal{L}_{TPM} = \mathcal{L}_{trpl, 1} + \mathcal{L}_{trpl, 2} + \mathcal{L}_{trpl, 3}. \label{eq:l_TPM-CL}
\end{equation}


\subsection{Objective Function}\label{subsec:obj}
Given a batch of $B$ video-text pairs, the model generates a $B \times B$ similarity matrix. We employ the \textit{Negative-aware InfoNCE} loss $\mathcal{L}_{NegNCE}$ to jointly incorporate the effects of positive and hard negative pairs. Moreover, we also utilize the \textit{Triplet Partial Margin} loss $\mathcal{L}_{TPM}$ to model different levels of semantic similarity among triplet samples. Hence, the total training loss $\mathcal{L}_{all}$ is defined as:
\begin{equation}
     \mathcal{L}_{all}  = \mathcal{L}_{NegNCE}+\mathcal{L}_{TPM}.
     \label{eq:loss_all}
\end{equation}
\begin{table*}[]
    \centering
    \caption{Retrieval results on MSR-VTT-1kA. 
    $\dag$ denotes that results are obtained by our re-training.}
    \label{tab:res_msrvtt}
    \renewcommand{\arraystretch}{0.8}
    \setlength\tabcolsep{2pt}
    \scalebox{1.0}{
    \begin{tabular}{l| c c c c c c| c c c c c c}
        \toprule
       \multicolumn{1}{c}{} & \multicolumn{6}{|c}{Text-to-Video Retrieval} & \multicolumn{6}{|c}{Video-to-Text Retrieval} \\ \toprule
        {Method} & {R$@$1$\uparrow$} & {R$@$5$\uparrow$} & {R$@$10$\uparrow$} & {MdR$\downarrow$} & {MeanR$\downarrow$} & {rsum$\uparrow$} & {R$@$1$\uparrow$} & {R$@$5$\uparrow$} & {R$@$10$\uparrow$} & {MdR$\downarrow$} & {MeanR$\downarrow$} & {rsum$\uparrow$} \\ \toprule
       \multicolumn{13}{l}{\textit{CLIP-ViT-B/32}} \\ 
        {CLIP4Clip~\cite{clip4clip2021}} &44.5	&71.4	&81.6	&2.0	&15.3	&197.5 &- &- &- &-&- &- \\ 
        {CenterCLIP~\cite{centerclip2022}} & 44.2	&71.6	&82.1	&2.0	&15.1	&197.9 &42.8	&71.7	&82.2	&2.0	&10.9	&196.7 \\ 
        {CLIP2TV~\cite{clip2tv2021}} & 46.1	&72.5	&82.9	&2.0	&15.2	&201.5	&43.9	&73	&82.8	&2.0	&11.1	&199.7 \\ 
        {XPool~\cite{xpool2022}} &46.9	&72.8	&82.2	&2.0	&14.3	&201.9 &- &- &- &- &- &- \\  
        {XCLIP$^\dag$~\cite{xclip2022}} & 47.4& 73.4& 83.1& 2.0& 13.7& 203.9 & 46.7&  72.7&  83.0&  2.0&  10.0&  202.4 \\   
        {DRL$^\dag$~\cite{drl2022}} &47.5& 73.8& 83.6& 2.0& 13.3& 204.9 &46.3& 72.7& 82.5& 2.0& 9.5& 201.5\\ 
        {TS2-Net$^\dag$~\cite{ts2net2022}} &47.2	&73.7	&83.1	&2.0	&13.1	&204.0 &44.8	&74.3	&84.0	&2.0	&9.3	&203.1  \\ 
        {Baseline} &45.3	&74.2	&83.5	&2.0	&13.0	&203.0	&45.5	&73.1	&83.9	&2.0	&9.6	&202.5  \\ 
        {Ours$_{ti}$} & 46.6& \textbf{75.0}& \textbf{84.1}& \textbf{2.0}&13.3& \textbf{205.7} & 46.0& \textbf{74.7}& 83.0& \textbf{2.0}& 9.5& \textbf{203.7}\\
        {Ours$_{wti}$} &46.9& \textbf{74.6}& \textbf{84.2}& \textbf{2.0}& \textbf{12.8}& \textbf{205.7}	&46.2& 73.7& \textbf{84.2}& \textbf{2.0}& \textbf{8.8}& \textbf{204.1}  \\ 
         
        \midrule
        \multicolumn{12}{l}{\textit{CLIP-ViT-B/16}} \\  
         {CenterCLIP~\cite{centerclip2022}} & 48.4	& 73.8	& 82.0	& 2.0	& 13.8	& 204.2	& 47.7	& 75.0	& 83.3	& 2.0	& 10.2	& 206.0  \\
         {CLIP2TV~\cite{clip2tv2021}} &49.3	&74.7	&83.6	&2.0	&13.5	&207.6	&46.9	&75.0	&85.1	&2.0	&10.0	&207.0  \\
       {DRL$^\dag$~\cite{drl2022}} &49.4& 76.4& 84.2& 2.0& 13.2& 210.0 &47.0& 77.1& 84.4& 2.0& 9.2& 208.5 \\
       {XCLIP$^\dag$~\cite{xclip2022}} &49.0& 76.9& 83.7& 2.0& 13.6& 209.6 &47.9& 75.0& 83.2& 2.0& 9.8& 206.1\\
       {TS2-Net$^\dag$~\cite{ts2net2022}} &47.8& 76.8& 85.2& 2.0& 13.7& 209.8 & 47.8	& 76.0	& 84.6	& 2.0	& 8.5	& 208.4\\
       {Baseline} &48.6 &74.8	&84.4	&2.0 &13.6	&207.8	&\textbf{48.0}	&75.9	&83.1	&2.0	&9.6 &207.0 \\ 
       {Ours$_{ti}$} &  \textbf{49.3}  &\textbf{77.0}  &\textbf{85.9}  &\textbf{2.0}	 &\textbf{12.7}	 &\textbf{212.2}  &\textbf{47.9}	 &76.0	 &\textbf{85.1}	 &\textbf{2.0}	 &9.1	 &\textbf{209.0} \\ 
       {Ours$_{wti}$} & \textbf{49.9}& 75.8& \textbf{85.5}& \textbf{2.0} & \textbf{12.5}& \textbf{211.2} &\textbf{49.6}& \textbf{76.3}& \textbf{85.0}&\textbf{2.0}& \textbf{8.5}& \textbf{210.9} \\ 
       \bottomrule
     \end{tabular} }
\end{table*}

\begin{table}[]
    \centering
    \caption{Retrieval results on MSVD. 
    $\dag$ denotes re-training.}%
    \label{tab:res_msvd}
    \renewcommand{\arraystretch}{0.8}
    \setlength\tabcolsep{2pt}
    \begin{tabular}{l| c c c c c c}
        \toprule
         {Method} & {R$@$1$\uparrow$} & {R$@$5$\uparrow$} & {R$@$10$\uparrow$} & {MdR$\downarrow$} & {MeanR$\downarrow$} & {rsum$\uparrow$} \\ \toprule
        {CLIP4Clip~\cite{clip4clip2021}} & 45.2	&75.5	&84.3	&2.0	&10.3	&205.0 \\ 
        {CLIP2TV~\cite{clip2tv2021}} & 47.0	&76.5	&85.1	&2.0	&10.1	&208.6 \\ 
        {DRL$^\dag$~\cite{drl2022}} & 46.5&  76.3&  85.0&  2.0& 10.7& 207.8 \\ 
        {TS2-Net$^\dag$~\cite{ts2net2022}} & 44.0	&75.5	&84.6	&2.0	&10.4	&204.1 \\
        {Baseline} & 44.0	&75.2	&84.2	&2.0	&10.9	&203.4 \\
        {Ours$_{ti}$} & 46.1 	& \textbf{76.4} & 	\textbf{85.0} & 	\textbf{2.0} & 	\textbf{10.1} & 207.5 \\
         
        {Ours$_{wti}$} & \textbf{46.9}&  \textbf{76.8}&  \textbf{85.6}&  \textbf{2.0}&  \textbf{9.7}&  \textbf{209.3} \\ 
\bottomrule
    \end{tabular}
\end{table}

\begin{table}[]
    \centering
    \caption{Retrieval results on DiDeMo. 
    $\dag$ denotes re-training.}
    \label{tab:res_didemo}
    \renewcommand{\arraystretch}{0.8}
    \setlength\tabcolsep{2pt}
    \begin{tabular}{l| c c c c c c}
        \toprule
         {Method} & {R$@$1$\uparrow$} & {R$@$5$\uparrow$} & {R$@$10$\uparrow$} & {MdR$\downarrow$} & {MeanR$\downarrow$} & {rsum$\uparrow$} \\ \toprule
        {CLIP4Clip~\cite{clip4clip2021}} & 42.5	&70.2	&80.6	&2.0	&17.5	&193.3 \\ 
        {CLIP2TV~\cite{clip2tv2021}} & 45.5	&69.7	&80.6	&2.0	&17.1	&195.8 \\ 
        {TS2-Net$^\dag$~\cite{ts2net2022}} & 41.5& 70.9& 80.6& 2.0& 13.9& 193.0 \\ 
        {DRL$^\dag$~\cite{drl2022}} & 46.5& 73.9& 83.5& 2.0& 13.3& 203.9 \\ 
        {Baseline} &  44.4	&73.3	&82.6	&2.0	&13.1	&200.3 \\ 
        {Ours$_{ti}$} & 45.2	&\textbf{74.1}	&\textbf{84.3}	&\textbf{2.0}	&\textbf{12.7}	&\textbf{203.6} \\
        {Ours$_{wti}$} &\textbf{46.7}	&\textbf{75.6}	&\textbf{84.0}	&\textbf{2.0}	&\textbf{11.7}	&\textbf{206.3}  \\ 
\bottomrule
    \end{tabular}
\end{table}

\begin{table}[]
    \centering
    \caption{Retrieval results on ActivityNet. 
    $\dag$ denotes re-training.}
    \label{tab:res_activity}
    \renewcommand{\arraystretch}{0.8}
    \setlength\tabcolsep{2pt}
    \begin{tabular}{l| c c c c c c}
        \toprule
         {Method} & {R$@$1$\uparrow$} & {R$@$5$\uparrow$} & {R$@$10$\uparrow$} & {MdR$\downarrow$} & {MeanR$\downarrow$} & {rsum$\uparrow$} \\ \toprule
        {CLIP4Clip~\cite{clip4clip2021}} &40.5	&72.4	&-	&2.0	&7.5	&-  \\  
        {CenterCLIP~\cite{centerclip2022}} &43.9 	&75.3 	&85.2 	&2.0 	&7.0 	&204.4 \\
        {TS2-Net$^\dag$~\cite{ts2net2022}} & 39.9&  72.3&  84.3&  2.0&  8.5&  196.5 \\ 
        {DRL~\cite{drl2022}} & 44.2	& 74.5	& 86.1	& 2.0	& - & 204.8 \\     
        {Baseline} &41.1& 72.3&84.1& 2.0& 8.2& 197.5 \\  
        {Ours$_{ti}$} &\textbf{44.8}& 74.4& 85.1& \textbf{2.0}& 7.4& \textbf{204.3}   \\  
        {Ours$_{wti}$} &\textbf{44.9}& \textbf{76.1}& \textbf{86.5}& \textbf{2.0}& \textbf{6.6}& \textbf{207.5}  \\ 
        \bottomrule
    \end{tabular}
  \vspace{-0.2cm}
\end{table}

\section{Experiments}\label{sec:exp}
\subsection{Experimental Setting}\label{subsec:exp_setting}
\textbf{Datasets}
To validate the effectiveness, we conduct experiments on four popular text-video retrieval datasets, including MSR-VTT~\cite{msrvtt2016}, MSVD~\cite{msvd2011}, DiDeMo~\cite{didemo2017} and ActivityNet~\cite{activitynet2015}.
\textbf{MSR-VTT}~\cite{msrvtt2016} is a general video dataset collected from YouTube and contains 10k videos and 200k captions. The videos range in length from 10 to 32 seconds. We train models on 9K videos, and report results on the 1K-A test set like~\cite{ts2net2022,xclip2022,drl2022}. 
\textbf{MSVD}~\cite{msvd2011} contains 1,970 videos and the duration of videos varies from 1 to 62 seconds. There are 40 English captions annotated for each video. The number of videos in the train/validation/test split is 1,200/100/670, respectively.
\textbf{DiDeMo}~\cite{didemo2017} is one of the largest and most diverse datasets for the temporal localization of events in videos given natural language descriptions and contains 10k Flickr videos annotated with 40k sentences. Following earlier studies~\cite{clip4clip2021,ts2net2022,xclip2022}, all captions from a video are concatenated together for video-paragraph retrieval.
\textbf{ActivityNet}~\cite{activitynet2015} contains 20k YouTube videos with 100k caption annotations. 
The videos are 120 seconds long on average. 
We concatenate all of a video's descriptions into one paragraph and evaluate the model with video-paragraph retrieval on the \textit{val1} split.



\textbf{Evaluation Metrics}
For a fair comparison, we evaluate the experimental results using standard text-video retrieval metrics: Recall at Rank K (R@K, higher is better), Median Rank (MdR, lower is better), Mean Rank (MeanR, lower is better) and rsum (higher is better). R@K calculates the percentage of correct samples in the top-K retrieved points to the query sample. 
Following previous works~\cite{clip2tv2021,ts2net2022,xclip2022}, we report results for R@1, R@5, R@10. In order to reflect the overall retrieval performance, we also sum together all the R@K results as rsum like in~\cite{Chen2020FineGrainedVR,Dong2021DualEF,hanet2021,ts2net2022}, which is the main concern in our experiments.
MdR measures the median rank of correct items in the retrieved ranking list and MeanR calculates the mean rank of correct items in the retrieved ranking list. 

\textbf{Implementation Details}
Our experiments are conducted on 8 NVIDIA Tesla V100 32GB GPUs using PyTorch. We initialize the text and video encoder with pre-trained weights from CLIP~\cite{clip2021}, while other modules are initialized randomly. We adopt the Adam optimizer~\cite{Kingma2014AdamAM} to train our model and decay the learning rate using a cosine schedule strategy~\cite{Loshchilov2016SGDRSG}. We set the learning rate  1e-7 and 1e-4 for text/video encoder and other modules, respectively. For MSR-VTT and MSVD, we set the max query text length, max video frame length and batch size to 32, 12 and 128, and apply $\mathbf{v}_{tokens}$ for the temporal encoder. We set the max query text length and max video frame length as 64 in ActivityNet and DiDeMo. Because of GPU memory limitations, we reduce the batch size of DiDeMo and ActivityNet to 64 and adopt $\mathbf{f}_{cls}$ for the temporal encoder.
We perform ablation experiments on the MSR-VTT dataset and the base model is ViT-B/16, while for the other datasets, the base model is ViT-B/32.
During training, we set the NegNCE loss weight $\gamma_1=1.0$ and $\gamma_2=0.5$ (in Eq.~\ref{eq:pos_neg}), and the TPM-CL parameters $\tau=0.6$ and $\delta=0.6$(in Eq.~\ref{eq:triplet_mask} and Eq.~\ref{eq:triplet_loss}).

\vspace{-0.2cm}
\subsection{Comparison with State-of-the-Art Methods}
We compare our approach against recent works (CLIP4CLIP, TS2-Net, DRL, \textit{etc}.) on MSR-VTT, MSVD, DiDeMo and ActivityNet datasets. 
Note that the performance may be affected by many factors, such as environment and algorithm module settings.
To mitigate the influence of the environment (\textit{e.g.}, GPU memory and version), we re-trained some experiments of previous methods in a unified environment setting. 
For a fair comparison with different methods, we show the results of our approach with two similarity calculation methods, \emph{i.e.}, \textbf{TI} and \textbf{WTI} \cite{drl2022}(denoted as $Ours_{ti}$ and $Ours_{wti}$ in Tab.~\ref{tab:res_msrvtt}-\ref{tab:res_activity}).
We set our baseline model as the degraded model, which removes the two core modules (DMAE and TPM-CL) and applies TI for similarity calculation.

We can see that our approach notably outperforms the baseline model in terms of all evaluation metrics and achieves the state-of-the-art performance. 
For the MSR-VTT dataset in Tab.\ref{tab:res_msrvtt}, our approach outperforms existing methods by a large margin on both ViT-B/32 and ViT-B/16 at two retrieval directions. 
Specifically, 
$Ours_{wti}$ outperforms DRL by nearly 1\% and over 2\% improvements on rsum of ViT-B/32 at two directions, respectively.
When compared with the baseline model using ViT-B/16, $Ours_{ti}$ obtains \textbf{4.4\%} and \textbf{2.0\%} improvements at two directions, respectively, while $Ours_{wti}$ largely improves rsum by \textbf{3.4\%} and \textbf{3.9\%}, where the R@1 gains 1.3\% and 1.6\% improvement. Moreover, compared to the previous SOTA methods (\emph{i.e.}, DRL and TS2-Net) using ViT-B/16, we have also over 2\% improvements on rsum at two directions.

We also further verify the generalization and robustness of our approach on MSVD, DiDeMo and ActivityNet.
Precisely, on the MSVD dataset as shown in Tab.~\ref{tab:res_msvd}, we observe that $Ours_{ti}$ outperforms the baseline model by \textbf{4.1\%} improvement on rsum, while $Ours_{wti}$ achieves 1.5\% gains compared to DRL.
For the DiDeMo dataset in Tab.~\ref{tab:res_didemo}, compared with the baseline and DRL, $Ours_{wti}$ surpasses all their evaluation performance and gains improvements of \textbf{6.0\%} and 2.4\% on rsum, respectively.
In the case of the ActivityNet dataset in Tab.~\ref{tab:res_activity}, our approach outperforms other existing methods by a large margin and achieves SOTA results on all evaluation metrics.
In general, the steady progress on several benchmarks is a solid indication of the effectiveness of our approach.


\vspace{-0.1cm}
\subsection{Ablation Study}\label{subsec:ablation}
In this section, we conduct ablation experiments on the MSR-VTT to verify the effectiveness of each module in our approach. 
\begin{table}[]
    \centering
    \caption{Retrieval performance with different settings of DMAE on the MSR-VTT.}
    \label{tab:res_c1}
    \renewcommand{\arraystretch}{0.8}
    \setlength\tabcolsep{1pt}
    \begin{tabular}{l| c c c c c c}
        \toprule
        {Method} & {R$@$1$\uparrow$} & {R$@$5$\uparrow$} & {R$@$10$\uparrow$} & {MdR$\downarrow$} & {MeanR$\downarrow$} & {rsum$\uparrow$} \\ \toprule
        {Baseline} &48.6 &74.8	&84.4	&2.0 &13.6	&207.8 \\ \midrule
        {Exp1(+NegNCE)} & 49.3	& 75.9& 83.8& 2.0	&12.8	& 209.0 \\ 
        {Exp2(+NegNCE+TA)} & 48.7 &76.5 &84.3 &2.0 &\textbf{12.7}	& \textbf{209.5} \\
        {Exp3(+NegNCE+VA)} & \textbf{50.0} &75.7 &84.5 &\textbf{1.5} &\textbf{12.7}	& \textbf{210.2} \\ 
        {Exp4(+All)} &49.5 & \textbf{76.7} & \textbf{84.7} &2.0	&12.8	&\textbf{210.9}  \\ \bottomrule
    \end{tabular}
\end{table}

\begin{table}[]
    \centering
    \caption{Ablation studies about the weighting parameters of NegNCE loss (in Eq.~\ref{eq:pos_neg}) on the MSR-VTT. The experiment setting is the same as Exp4 in Tab.\ref{tab:res_c1}.}.
    \label{tab:res_c1_weight}
    \renewcommand{\arraystretch}{0.8}
    \setlength\tabcolsep{2pt}
    \begin{tabular}{l| c c c c c c}
        \toprule
        {Method} & {R$@$1$\uparrow$} & {R$@$5$\uparrow$} & {R$@$10$\uparrow$} & {MdR$\downarrow$} & {MeanR$\downarrow$} & {rsum$\uparrow$} \\ \toprule
        {Baseline} &48.6 &74.8	&84.4	&\textbf{2.0} &13.6	&207.8 \\ \midrule
        \multicolumn{7}{l}{\textit{with $\gamma_1=1.0$ by default}} \\  
        {$\gamma_2=0.0$} & 49.2	&76.0	&84.4	&\textbf{2.0}	&13.0	&209.6	\\
        {$\gamma_2=0.3$} & \textbf{50.1}	&75.4	&84.4	&\textbf{2.0}	&13.0	&209.9	\\
        {$\gamma_2=0.5$} &49.5 &\textbf{76.7} & 84.7 &\textbf{2.0}	&\textbf{12.8}	&\textbf{210.9} \\ 
        {$\gamma_2=0.7$} & 48.4 &75.7 &\textbf{85.5} &\textbf{2.0} &13.7 &209.6 \\ \bottomrule
    \end{tabular}
  \vspace{-0.3cm}
\end{table}

\subsubsection{Effectiveness of Dual-Modal Attention-Enhanced Module}
We first investigate the impact of DMAE and conduct an ablation study to compare different variants of each component. As shown in Tab.~\ref{tab:res_c1}, all variants see a big boost in terms of retrieval performance. 
Specifically, compared with the baseline model, 
DMAE with only NegNCE (\emph{i.e.}, Exp1) achieves merely 1.2\% gains on rsum. When DMAE is equipped with the components of \textit{Textual Attention} and \textit{Visual Attention} (\emph{i.e.}, Exp2 and Exp3), the performance further improves by 1.7\% and 2.4\%, respectively. At last, DMAE with all components (\emph{i.e.}, Exp4) obtains a notable improvement of \textbf{3.1\%} on rsum, where the R@1, R@5, R@10 gain 0.9\%, 1.9\%, 0.3\% improvements, respectively. Therefore, we conclude that all components in DAME contribute to the retrieval task and different components can promote each other to achieve better results.


\subsubsection{The Impact of weight in NegNCE loss}
To explore the impact of different weights in the NegNCE loss, we also design a group of experiments by setting different weighting parameters $\gamma_2$ with a fixed setting of $\gamma_1=1.0$ (\emph{i.e.}, the original InfoNCE loss is a special case of the NegNCE loss if $\gamma_2=0.0$). From Tab.~\ref{tab:res_c1_weight}, it can be seen that the overall retrieval performance initially increases before reaching saturation (\emph{i.e.}, $\gamma_2=0.5$), and then declines slightly. The main reason may be that when $\gamma_2$ is large, the model weights too much on the hard negative pairs. Conversely, if the $\gamma_2$ is small, the effect of hard negative pairs may be underestimated.

\begin{table}[!htb]
    \centering
    \caption{Retrieval performance with TPM-CL on the MSR-VTT.}
    \label{tab:res_c1c2}
    \renewcommand{\arraystretch}{0.8}
    \setlength\tabcolsep{2pt}
    \begin{tabular}{l| c c c c c c}
        \toprule
        {Method} & {R$@$1$\uparrow$} & {R$@$5$\uparrow$} & {R$@$10$\uparrow$} & {MdR$\downarrow$} & {MeanR$\downarrow$} & {rsum$\uparrow$} \\ \toprule
        {Baseline} &48.6 &74.8	&84.4	&\textbf{2.0} &13.6	&207.8 \\ 
        {+TPM-CL} & \textbf{49.4} &76.1 &{85.1} &\textbf{2.0}	&13.2	&\textbf{210.6} \\ 
        {+DMAE+TPM-CL} & 49.3 &\textbf{77.0} &\textbf{85.9} &\textbf{2.0}	&\textbf{12.7}	&\textbf{212.2} \\ \bottomrule
    \end{tabular}
  \vspace{-0.2cm}
\end{table}

\begin{table}[]
    \centering
    \caption{Ablation studies about the hyper parameters of TPM-CL (in Eq.~\ref{eq:triplet_mask} and Eq.~\ref{eq:triplet_loss}) on the MSR-VTT.}.
    \label{tab:hyper_TPM-CL}
    \renewcommand{\arraystretch}{0.8}
    \setlength\tabcolsep{2pt}
    \begin{tabular}{l| c c c c c c}
        \toprule
        {Method} & {R$@$1$\uparrow$} & {R$@$5$\uparrow$} & {R$@$10$\uparrow$} & {MdR$\downarrow$} & {MeanR$\downarrow$} & {rsum$\uparrow$} \\ 
        \toprule
        {Baseline} &48.6 &74.8	&84.4	&\textbf{2.0} &13.6	&207.8 \\ 
        \midrule
        \multicolumn{7}{l}{\textit{with triplet ranking loss margin $\delta=0.2$}} \\  
        {$\tau=0.2$} &47.9	&76.1 &84.6	&\textbf{2.0}	&13.0	&208.6 \\
        {$\tau=0.6$} &48.8  &\textbf{76.3} &\textbf{85.0} &\textbf{2.0}    &12.8   &\textbf{210.1} \\  
        {$\tau=0.9$} &\textbf{49.2}  &75.6 &84.5 &\textbf{2.0}    &13.6   &209.3 \\  
        \midrule
        \multicolumn{7}{l}{\textit{with masked feature ratio $\tau=0.6$}} \\  
        {$\delta=0.2$} &48.8	&76.3	&85.0	&\textbf{2.0}	&12.8	&210.1 \\
        {$\delta=0.6$} &\textbf{49.3} &\textbf{77.0} &\textbf{85.9} &\textbf{2.0} &\textbf{12.7}	&\textbf{212.2} \\ 
        {$\delta=1.0$} &48.4    &76.6   &84.6   &\textbf{2.0}   &12.9   &209.6 \\ 
        \bottomrule
    \end{tabular}
\end{table}

\subsubsection{Effectiveness of Triplet Partial Margin Contrastive Module}
 Similarly, we also perform experiments to validate the effect of TPM-CL. The results in Tab.\ref{tab:res_c1c2} clearly demonstrate that the model with only TPM-CL outperforms the baseline model by 2.8\% on rsum, while the full model equipped with DMAE and TPM-CL further obtains a large margin of \textbf{4.4\%}, indicating that our two core modules are both beneficial to improve the retrieval performance.

\subsubsection{The Impact of hyper parameters in TPM-CL}
We conduct a group of experiments with different values of the triplet ranking loss margin $\delta$ and the masked feature ratio $\tau$. From Tab.~\ref{tab:hyper_TPM-CL}, we get the best rsum performance when $\tau=0.6$ for a fixed setting of $\delta=0.2$.  
Furthermore, with $\delta$ varying, the overall performance first improves from 210.1 to 212.2 and then declines if $\tau$ is fixed to 0.6. 
The main reason may be that a large $\tau$ makes the masked sample quite different from the original one and a large $\delta$ means the difference between them should be large enough to make sense. 
Thus, a moderate setting of both $\tau$ and $\delta$ can encourage the learning to examine the fine-grained semantic similarity among triplets.

\subsection{Qualitative Analysis}

\begin{figure}[] 
  \centering
  \includegraphics[width=0.9\linewidth]{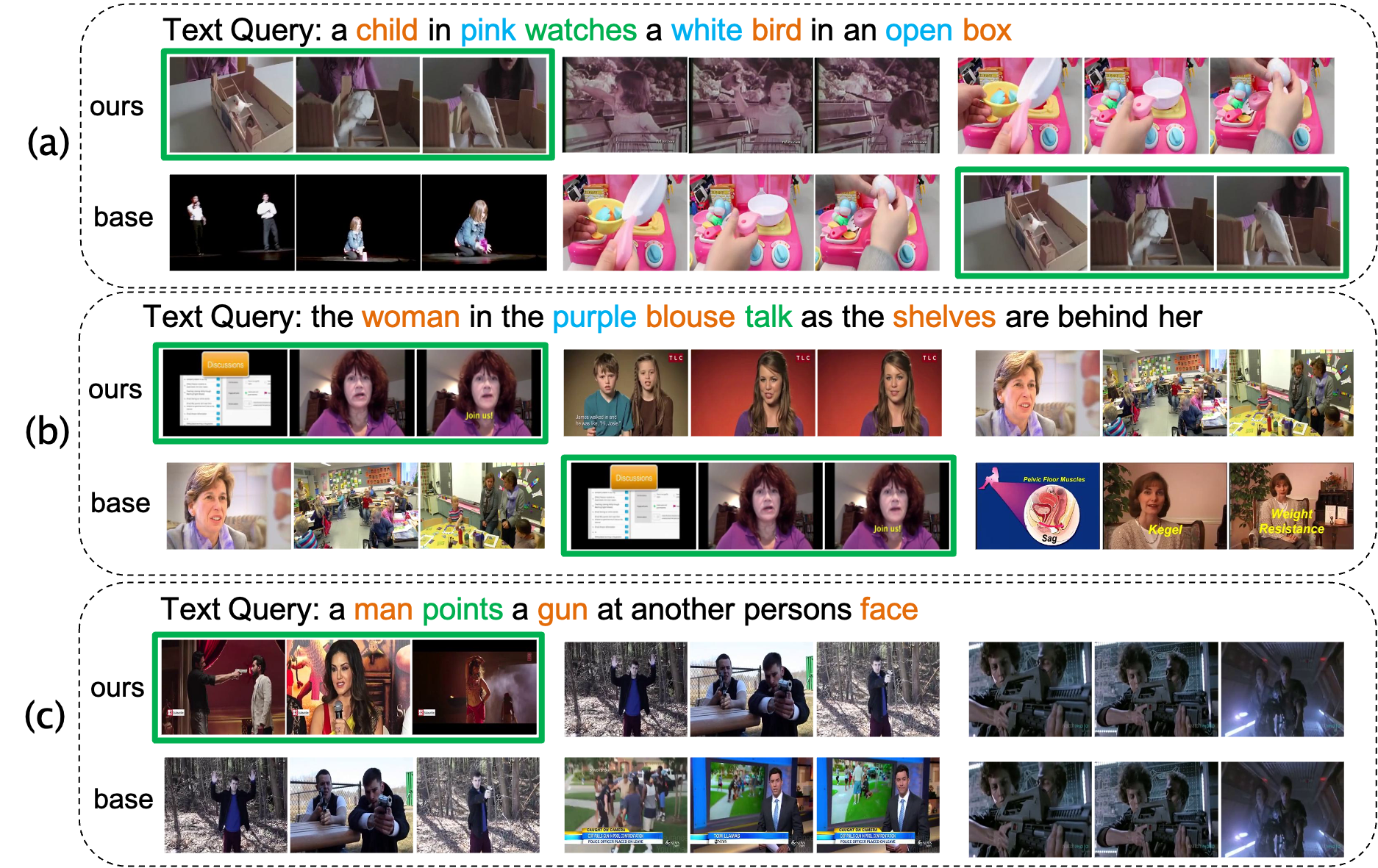}
  \caption{Visualization of text-to-video retrieval results on MSR-VTT. For each query, the top-3 results are displayed and sorted based on their similarity scores. The upper half of the two retrieval groups are the results with our full model, while the lower half are the retrieval results with the baseline model. Green box: ground truth.}
  \label{fig:examples}
  \vspace{-0.2cm}
\end{figure}

To qualitatively validate the effectiveness of our approach, we visualize some text-to-video retrieval examples from the MSR-VTT in Fig.~\ref{fig:examples}. Specifically, as in Fig.~\ref{fig:examples}(a) and (b), our model retrieves the correct videos that contain all matched fragments described in the text query (\emph{i.e.}, \textit{"child in pink"}, \textit{"a white bird"} and \textit{"an open box"} in (a), \textit{"woman in the purple blouse"} and \textit{"shelves"} in (b), respectively). 
Our model can successfully differentiate positives from those hard negatives with partly matched fragments. 
The examples in Fig.~\ref{fig:examples}(c) show that our model is capable of accurately capturing the corresponding hard positive video, even if the matched fragment is only present in a small segment of the video.
Meanwhile, we find that our model can focus on the relevant videos with fully or partly matched fragments while eliminating the false positives without matched fragments (\textit{e.g.}, the 2nd example in the lower part of Fig.~\ref{fig:examples}(c)).
To summarize, our model can accurately capture the correct videos and retrieve more related videos compared to the baseline model, demonstrating the merits of our approach.

\section{Conclusion}
This paper proposed a novel \textit{Dual-Modal Attention-Enhanced module (DMAE)} to mine hard negatives from textual and visual clues, and introduced a \textit{Negative-aware InfoNCE (NegNCE)} loss to adaptively incorporate them into the training objective.
Then we presented a new \textit{Triplet Partial Margin Contrastive Learning (TPM-CL)} module, which aims to focus on the automatic constitution of triplet samples and capture the fine-grained semantic similarity among them.
The effectiveness and superiority of our proposed method have been clearly demonstrated in comprehensive experiments on four text-video retrieval benchmarks.




\bibliographystyle{ACM-Reference-Format}
\balance
\bibliography{sample_t2v}


\begin{thebibliography}{57}


\ifx \showCODEN    \undefined \def \showCODEN     #1{\unskip}     \fi
\ifx \showDOI      \undefined \def \showDOI       #1{#1}\fi
\ifx \showISBNx    \undefined \def \showISBNx     #1{\unskip}     \fi
\ifx \showISBNxiii \undefined \def \showISBNxiii  #1{\unskip}     \fi
\ifx \showISSN     \undefined \def \showISSN      #1{\unskip}     \fi
\ifx \showLCCN     \undefined \def \showLCCN      #1{\unskip}     \fi
\ifx \shownote     \undefined \def \shownote      #1{#1}          \fi
\ifx \showarticletitle \undefined \def \showarticletitle #1{#1}   \fi
\ifx \showURL      \undefined \def \showURL       {\relax}        \fi
\providecommand\bibfield[2]{#2}
\providecommand\bibinfo[2]{#2}
\providecommand\natexlab[1]{#1}
\providecommand\showeprint[2][]{arXiv:#2}

\bibitem[Aizawa(2003)]%
        {Aizawa2003AnIP}
\bibfield{author}{\bibinfo{person}{Akiko Aizawa}.}
  \bibinfo{year}{2003}\natexlab{}.
\newblock \showarticletitle{An information-theoretic perspective of tf-idf
  measures}.
\newblock \bibinfo{journal}{\emph{Inf. Process. Manag.}}  \bibinfo{volume}{39}
  (\bibinfo{year}{2003}), \bibinfo{pages}{45--65}.
\newblock


\bibitem[Bain et~al\mbox{.}(2021)]%
        {frozen2021}
\bibfield{author}{\bibinfo{person}{Max Bain}, \bibinfo{person}{Arsha Nagrani},
  \bibinfo{person}{G{\"u}l Varol}, {and} \bibinfo{person}{Andrew Zisserman}.}
  \bibinfo{year}{2021}\natexlab{}.
\newblock \showarticletitle{Frozen in Time: A Joint Video and Image Encoder for
  End-to-End Retrieval}. In \bibinfo{booktitle}{\emph{2021 IEEE/CVF
  International Conference on Computer Vision (ICCV)}}.
  \bibinfo{pages}{1708--1718}.
\newblock


\bibitem[Bogolin et~al\mbox{.}(2021)]%
        {querybank2022}
\bibfield{author}{\bibinfo{person}{Simion-Vlad Bogolin}, \bibinfo{person}{Ioana
  Croitoru}, \bibinfo{person}{Hailin Jin}, \bibinfo{person}{Yang Liu}, {and}
  \bibinfo{person}{Samuel Albanie}.} \bibinfo{year}{2021}\natexlab{}.
\newblock \showarticletitle{Cross Modal Retrieval with Querybank
  Normalisation}. In \bibinfo{booktitle}{\emph{2022 IEEE/CVF Conference on
  Computer Vision and Pattern Recognition (CVPR)}}.
  \bibinfo{pages}{5184--5195}.
\newblock


\bibitem[Chen and Dolan(2011)]%
        {msvd2011}
\bibfield{author}{\bibinfo{person}{David~L. Chen} {and}
  \bibinfo{person}{William~B. Dolan}.} \bibinfo{year}{2011}\natexlab{}.
\newblock \showarticletitle{Collecting Highly Parallel Data for Paraphrase
  Evaluation}. In \bibinfo{booktitle}{\emph{Annual Meeting of the Association
  for Computational Linguistics}}.
\newblock


\bibitem[Chen et~al\mbox{.}(2020)]%
        {Chen2020FineGrainedVR}
\bibfield{author}{\bibinfo{person}{Shizhe Chen}, \bibinfo{person}{Yida Zhao},
  \bibinfo{person}{Qin Jin}, {and} \bibinfo{person}{Qi Wu}.}
  \bibinfo{year}{2020}\natexlab{}.
\newblock \showarticletitle{Fine-Grained Video-Text Retrieval With Hierarchical
  Graph Reasoning}. In \bibinfo{booktitle}{\emph{2020 IEEE/CVF Conference on
  Computer Vision and Pattern Recognition (CVPR)}}.
  \bibinfo{pages}{10635--10644}.
\newblock


\bibitem[Cheng et~al\mbox{.}(2021)]%
        {camoe2021}
\bibfield{author}{\bibinfo{person}{Xingyi Cheng}, \bibinfo{person}{Hezheng
  Lin}, \bibinfo{person}{Xiangyu Wu}, \bibinfo{person}{F. Yang}, {and}
  \bibinfo{person}{Dong Shen}.} \bibinfo{year}{2021}\natexlab{}.
\newblock \showarticletitle{Improving Video-Text Retrieval by Multi-Stream
  Corpus Alignment and Dual Softmax Loss}.
\newblock \bibinfo{journal}{\emph{ArXiv}}  \bibinfo{volume}{abs/2109.04290}
  (\bibinfo{year}{2021}).
\newblock


\bibitem[Dong et~al\mbox{.}(2021)]%
        {Dong2021DualEF}
\bibfield{author}{\bibinfo{person}{Jianfeng Dong}, \bibinfo{person}{Xirong Li},
  \bibinfo{person}{Chaoxi Xu}, \bibinfo{person}{Xun Yang},
  \bibinfo{person}{Gang Yang}, \bibinfo{person}{Xun Wang}, {and}
  \bibinfo{person}{Meng Wang}.} \bibinfo{year}{2021}\natexlab{}.
\newblock \showarticletitle{Dual Encoding for Video Retrieval by Text}.
\newblock \bibinfo{journal}{\emph{IEEE Transactions on Pattern Analysis and
  Machine Intelligence (TPAMI)}}  \bibinfo{volume}{44} (\bibinfo{year}{2021}),
  \bibinfo{pages}{4065--4080}.
\newblock


\bibitem[Falcon et~al\mbox{.}(2022)]%
        {falcon_relevance-based_2022}
\bibfield{author}{\bibinfo{person}{Alex Falcon}, \bibinfo{person}{Swathikiran
  Sudhakaran}, \bibinfo{person}{Giuseppe Serra}, \bibinfo{person}{Sergio
  Escalera}, {and} \bibinfo{person}{Oswald Lanz}.}
  \bibinfo{year}{2022}\natexlab{}.
\newblock \showarticletitle{Relevance-based Margin for Contrastively-trained
  Video Retrieval Models}.
\newblock \bibinfo{journal}{\emph{Proceedings of the 2022 International
  Conference on Multimedia Retrieval (ICMR)}}, \bibinfo{pages}{146--157}.
\newblock


\bibitem[Fang et~al\mbox{.}(2021)]%
        {clip2video2021}
\bibfield{author}{\bibinfo{person}{Han Fang}, \bibinfo{person}{Pengfei Xiong},
  \bibinfo{person}{Luhui Xu}, {and} \bibinfo{person}{Yu Chen}.}
  \bibinfo{year}{2021}\natexlab{}.
\newblock \showarticletitle{CLIP2Video: Mastering Video-Text Retrieval via
  Image CLIP}.
\newblock \bibinfo{journal}{\emph{ArXiv}}  \bibinfo{volume}{abs/2106.11097}
  (\bibinfo{year}{2021}).
\newblock


\bibitem[Fang et~al\mbox{.}(2022)]%
        {acp2022}
\bibfield{author}{\bibinfo{person}{Sheng Fang}, \bibinfo{person}{Shuhui Wang},
  \bibinfo{person}{Junbao Zhuo}, \bibinfo{person}{Qingming Huang},
  \bibinfo{person}{Bin Ma}, \bibinfo{person}{Xiaoming Wei}, {and}
  \bibinfo{person}{Xiaolin Wei}.} \bibinfo{year}{2022}\natexlab{}.
\newblock \showarticletitle{Concept Propagation via Attentional Knowledge Graph
  Reasoning for Video-Text Retrieval}. In \bibinfo{booktitle}{\emph{Proceedings
  of the 30th ACM International Conference on Multimedia (ACM MM)}}.
  \bibinfo{pages}{4789–4800}.
\newblock


\bibitem[Gabeur et~al\mbox{.}(2020)]%
        {MMT}
\bibfield{author}{\bibinfo{person}{Valentin Gabeur}, \bibinfo{person}{Chen
  Sun}, \bibinfo{person}{Alahari Karteek}, {and} \bibinfo{person}{Cordelia
  Schmid}.} \bibinfo{year}{2020}\natexlab{}.
\newblock \showarticletitle{Multi-modal Transformer for Video Retrieval}. In
  \bibinfo{booktitle}{\emph{European Conference on Computer Vision (ECCV)}}.
\newblock


\bibitem[Gao and Lu(2022)]%
        {sstvlmss2022}
\bibfield{author}{\bibinfo{person}{Yizhao Gao} {and} \bibinfo{person}{Zhiwu
  Lu}.} \bibinfo{year}{2022}\natexlab{}.
\newblock \showarticletitle{SST-VLM: Sparse Sampling-Twice Inspired
  Video-Language Model}. In \bibinfo{booktitle}{\emph{Asian Conference on
  Computer Vision (ACCV)}}.
\newblock


\bibitem[Gao et~al\mbox{.}(2021)]%
        {clip2tv2021}
\bibfield{author}{\bibinfo{person}{Zijian Gao}, \bibinfo{person}{Jingyu Liu},
  \bibinfo{person}{Sheng Chen}, \bibinfo{person}{Dedan Chang},
  \bibinfo{person}{Hao Zhang}, {and} \bibinfo{person}{Jinwei Yuan}.}
  \bibinfo{year}{2021}\natexlab{}.
\newblock \showarticletitle{{CLIP2TV:} An Empirical Study on Transformer-based
  Methods for Video-Text Retrieval}.
\newblock \bibinfo{journal}{\emph{ArXiv}}  \bibinfo{volume}{abs/2111.05610}
  (\bibinfo{year}{2021}).
\newblock


\bibitem[Gorti et~al\mbox{.}(2022)]%
        {xpool2022}
\bibfield{author}{\bibinfo{person}{Satya~Krishna Gorti}, \bibinfo{person}{Noel
  Vouitsis}, \bibinfo{person}{Junwei Ma}, \bibinfo{person}{Keyvan Golestan},
  \bibinfo{person}{Maksims Volkovs}, \bibinfo{person}{Animesh Garg}, {and}
  \bibinfo{person}{Guangwei Yu}.} \bibinfo{year}{2022}\natexlab{}.
\newblock \showarticletitle{X-Pool: Cross-Modal Language-Video Attention for
  Text-Video Retrieval}. In \bibinfo{booktitle}{\emph{2022 IEEE/CVF Conference
  on Computer Vision and Pattern Recognition (CVPR)}}.
  \bibinfo{pages}{4996--5005}.
\newblock


\bibitem[Gutmann and Hyv{\"a}rinen(2010)]%
        {nce2010}
\bibfield{author}{\bibinfo{person}{Michael~U Gutmann} {and}
  \bibinfo{person}{Aapo Hyv{\"a}rinen}.} \bibinfo{year}{2010}\natexlab{}.
\newblock \showarticletitle{Noise-contrastive estimation: A new estimation
  principle for unnormalized statistical models}. In
  \bibinfo{booktitle}{\emph{International Conference on Artificial Intelligence
  and Statistics}}.
\newblock


\bibitem[He et~al\mbox{.}(2019)]%
        {moco2019}
\bibfield{author}{\bibinfo{person}{Kaiming He}, \bibinfo{person}{Haoqi Fan},
  \bibinfo{person}{Yuxin Wu}, \bibinfo{person}{Saining Xie}, {and}
  \bibinfo{person}{Ross~B. Girshick}.} \bibinfo{year}{2019}\natexlab{}.
\newblock \showarticletitle{Momentum Contrast for Unsupervised Visual
  Representation Learning}. In \bibinfo{booktitle}{\emph{2020 IEEE/CVF
  Conference on Computer Vision and Pattern Recognition (CVPR)}}.
  \bibinfo{pages}{9726--9735}.
\newblock


\bibitem[Heilbron et~al\mbox{.}(2015)]%
        {activitynet2015}
\bibfield{author}{\bibinfo{person}{Fabian~Caba Heilbron},
  \bibinfo{person}{Victor Escorcia}, \bibinfo{person}{Bernard Ghanem}, {and}
  \bibinfo{person}{Juan~Carlos Niebles}.} \bibinfo{year}{2015}\natexlab{}.
\newblock \showarticletitle{ActivityNet: A large-scale video benchmark for
  human activity understanding}. In \bibinfo{booktitle}{\emph{2015 IEEE
  Conference on Computer Vision and Pattern Recognition (CVPR)}}.
  \bibinfo{pages}{961--970}.
\newblock


\bibitem[Hendricks et~al\mbox{.}(2017)]%
        {didemo2017}
\bibfield{author}{\bibinfo{person}{Lisa~Anne Hendricks},
  \bibinfo{person}{Oliver Wang}, \bibinfo{person}{Eli Shechtman},
  \bibinfo{person}{Josef Sivic}, \bibinfo{person}{Trevor Darrell}, {and}
  \bibinfo{person}{Bryan~C. Russell}.} \bibinfo{year}{2017}\natexlab{}.
\newblock \showarticletitle{Localizing Moments in Video with Natural Language}.
  In \bibinfo{booktitle}{\emph{2017 IEEE International Conference on Computer
  Vision (ICCV)}}. \bibinfo{pages}{5804--5813}.
\newblock


\bibitem[Honnibal and Montani(2017)]%
        {spacy2}
\bibfield{author}{\bibinfo{person}{Matthew Honnibal} {and}
  \bibinfo{person}{Ines Montani}.} \bibinfo{year}{2017}\natexlab{}.
\newblock \bibinfo{title}{{spaCy 2}: Natural language understanding with
  {B}loom embeddings, convolutional neural networks and incremental parsing}.
  (\bibinfo{year}{2017}).
\newblock


\bibitem[Jones(2021)]%
        {SprckJones2021ASI}
\bibfield{author}{\bibinfo{person}{Karen~Sp{\"a}rck Jones}.}
  \bibinfo{year}{2021}\natexlab{}.
\newblock \showarticletitle{A statistical interpretation of term specificity
  and its application in retrieval}.
\newblock \bibinfo{journal}{\emph{J. Documentation}}  \bibinfo{volume}{60}
  (\bibinfo{year}{2021}), \bibinfo{pages}{493--502}.
\newblock


\bibitem[Kalantidis et~al\mbox{.}(2020)]%
        {MoCHi2020}
\bibfield{author}{\bibinfo{person}{Yannis Kalantidis},
  \bibinfo{person}{Mert~Bulent Sariyildiz}, \bibinfo{person}{Noe Pion},
  \bibinfo{person}{Philippe Weinzaepfel}, {and} \bibinfo{person}{Diane
  Larlus}.} \bibinfo{year}{2020}\natexlab{}.
\newblock \showarticletitle{Hard Negative Mixing for Contrastive Learning}. In
  \bibinfo{booktitle}{\emph{Advances in Neural Information Processing Systems
  (NeurIPS)}}. \bibinfo{publisher}{Curran Associates, Inc.}
\newblock


\bibitem[Kingma and Ba(2014)]%
        {Kingma2014AdamAM}
\bibfield{author}{\bibinfo{person}{Diederik~P. Kingma} {and}
  \bibinfo{person}{Jimmy Ba}.} \bibinfo{year}{2014}\natexlab{}.
\newblock \showarticletitle{Adam: A Method for Stochastic Optimization}.
\newblock \bibinfo{journal}{\emph{ArXiv}}  \bibinfo{volume}{abs/1412.6980}
  (\bibinfo{year}{2014}).
\newblock


\bibitem[Kolesnikov et~al\mbox{.}(2021)]%
        {Dosovitskiy2020AnII}
\bibfield{author}{\bibinfo{person}{Alexander Kolesnikov},
  \bibinfo{person}{Alexey Dosovitskiy}, \bibinfo{person}{Dirk Weissenborn},
  \bibinfo{person}{Georg Heigold}, \bibinfo{person}{Jakob Uszkoreit},
  \bibinfo{person}{Lucas Beyer}, \bibinfo{person}{Matthias Minderer},
  \bibinfo{person}{Mostafa Dehghani}, \bibinfo{person}{Neil Houlsby},
  \bibinfo{person}{Sylvain Gelly}, \bibinfo{person}{Thomas Unterthiner}, {and}
  \bibinfo{person}{Xiaohua Zhai}.} \bibinfo{year}{2021}\natexlab{}.
\newblock \showarticletitle{An Image is Worth 16x16 Words: Transformers for
  Image Recognition at Scale}. In \bibinfo{booktitle}{\emph{International
  Conference on Learning Representations (ICLR)}}.
\newblock


\bibitem[Li et~al\mbox{.}(2021)]%
        {Li2021AlignAP}
\bibfield{author}{\bibinfo{person}{Dongxu Li}, \bibinfo{person}{Junnan Li},
  \bibinfo{person}{Hongdong Li}, \bibinfo{person}{Juan~Carlos Niebles}, {and}
  \bibinfo{person}{Steven C.~H. Hoi}.} \bibinfo{year}{2021}\natexlab{}.
\newblock \showarticletitle{Align and Prompt: Video-and-Language Pre-training
  with Entity Prompts}. In \bibinfo{booktitle}{\emph{2022 IEEE/CVF Conference
  on Computer Vision and Pattern Recognition (CVPR)}}.
  \bibinfo{pages}{4943--4953}.
\newblock


\bibitem[Li et~al\mbox{.}(2020)]%
        {HERO}
\bibfield{author}{\bibinfo{person}{Linjie Li}, \bibinfo{person}{Yen-Chun Chen},
  \bibinfo{person}{Yu Cheng}, \bibinfo{person}{Zhe Gan},
  \bibinfo{person}{Licheng Yu}, {and} \bibinfo{person}{Jingjing Liu}.}
  \bibinfo{year}{2020}\natexlab{}.
\newblock \showarticletitle{Hero: Hierarchical Encoder for Video+Language
  Omni-representation Pre-training}. In \bibinfo{booktitle}{\emph{Proceedings
  of the 2020 Conference on Empirical Methods in Natural Language Processing
  (EMNLP)}}.
\newblock


\bibitem[Li et~al\mbox{.}(2022)]%
        {videoclip2022}
\bibfield{author}{\bibinfo{person}{Yikang Li}, \bibinfo{person}{Jenhao Hsiao},
  {and} \bibinfo{person}{Chiu~Man Ho}.} \bibinfo{year}{2022}\natexlab{}.
\newblock \showarticletitle{VideoCLIP: A Cross-Attention Model for Fast
  Video-Text Retrieval Task with Image CLIP}. In
  \bibinfo{booktitle}{\emph{Proceedings of the 2022 International Conference on
  Multimedia Retrieval (ICMR)}}.
\newblock


\bibitem[Liu and Ye(2019)]%
        {liu2019strong}
\bibfield{author}{\bibinfo{person}{Fangyu Liu} {and} \bibinfo{person}{Rongtian
  Ye}.} \bibinfo{year}{2019}\natexlab{}.
\newblock \showarticletitle{A strong and robust baseline for text-image
  matching}.
\newblock \bibinfo{journal}{\emph{arXiv preprint arXiv:1906.01205}}
  (\bibinfo{year}{2019}).
\newblock


\bibitem[Liu et~al\mbox{.}(2023)]%
        {liu2023revisiting}
\bibfield{author}{\bibinfo{person}{Ruyang Liu}, \bibinfo{person}{Jingjia
  Huang}, \bibinfo{person}{Ge Li}, \bibinfo{person}{Jiashi Feng},
  \bibinfo{person}{Xinglong Wu}, {and} \bibinfo{person}{Thomas~H Li}.}
  \bibinfo{year}{2023}\natexlab{}.
\newblock \showarticletitle{Revisiting Temporal Modeling for CLIP-based
  Image-to-Video Knowledge Transferring}.
\newblock \bibinfo{journal}{\emph{Proceedings of the IEEE/CVF Conference on
  Computer Vision and Pattern Recognition (CVPR)}}.
\newblock


\bibitem[Liu et~al\mbox{.}(2021a)]%
        {hit2021}
\bibfield{author}{\bibinfo{person}{Song Liu}, \bibinfo{person}{Haoqi Fan},
  \bibinfo{person}{Shengsheng Qian}, \bibinfo{person}{Yiru Chen},
  \bibinfo{person}{Wenkui Ding}, {and} \bibinfo{person}{Zhongyuan Wang}.}
  \bibinfo{year}{2021}\natexlab{a}.
\newblock \showarticletitle{HiT: Hierarchical Transformer with Momentum
  Contrast for Video-Text Retrieval}. In \bibinfo{booktitle}{\emph{2021
  IEEE/CVF International Conference on Computer Vision (ICCV)}}.
  \bibinfo{pages}{11895--11905}.
\newblock


\bibitem[Liu et~al\mbox{.}(2021b)]%
        {tupleinfonce2021}
\bibfield{author}{\bibinfo{person}{Yunze Liu}, \bibinfo{person}{Qingnan Fan},
  \bibinfo{person}{Shanghang Zhang}, \bibinfo{person}{Hao Dong},
  \bibinfo{person}{Thomas~A. Funkhouser}, {and} \bibinfo{person}{Li Yi}.}
  \bibinfo{year}{2021}\natexlab{b}.
\newblock \showarticletitle{Contrastive Multimodal Fusion with TupleInfoNCE}.
  In \bibinfo{booktitle}{\emph{2021 IEEE/CVF International Conference on
  Computer Vision (ICCV)}}. \bibinfo{pages}{734--743}.
\newblock


\bibitem[Liu et~al\mbox{.}(2022)]%
        {ts2net2022}
\bibfield{author}{\bibinfo{person}{Yuqi Liu}, \bibinfo{person}{Pengfei Xiong},
  \bibinfo{person}{Luhui Xu}, \bibinfo{person}{Shengming Cao}, {and}
  \bibinfo{person}{Qin Jin}.} \bibinfo{year}{2022}\natexlab{}.
\newblock \showarticletitle{TS2-Net: Token Shift and Selection Transformer for
  Text-Video Retrieval}. In \bibinfo{booktitle}{\emph{Proceedings of the
  European Conference on Computer Vision (ECCV)}}.
\newblock


\bibitem[Loshchilov and Hutter(2017)]%
        {Loshchilov2016SGDRSG}
\bibfield{author}{\bibinfo{person}{Ilya Loshchilov} {and}
  \bibinfo{person}{Frank Hutter}.} \bibinfo{year}{2017}\natexlab{}.
\newblock \showarticletitle{SGDR: Stochastic Gradient Descent with Warm
  Restarts}. In \bibinfo{booktitle}{\emph{5th International Conference on
  Learning Representations (ICLR)}}.
\newblock


\bibitem[Lu et~al\mbox{.}(2022)]%
        {lgdn2022}
\bibfield{author}{\bibinfo{person}{Haoyu Lu}, \bibinfo{person}{Mingyu Ding},
  \bibinfo{person}{Nanyi Fei}, \bibinfo{person}{Yuqi Huo}, {and}
  \bibinfo{person}{Zhiwu Lu}.} \bibinfo{year}{2022}\natexlab{}.
\newblock \showarticletitle{{LGDN}: Language-Guided Denoising Network for
  Video-Language Modeling}. In \bibinfo{booktitle}{\emph{Advances in Neural
  Information Processing Systems (NeurIPS)}}.
\newblock


\bibitem[Luo et~al\mbox{.}(2020)]%
        {univl2020}
\bibfield{author}{\bibinfo{person}{Huaishao Luo}, \bibinfo{person}{Lei Ji},
  \bibinfo{person}{Botian Shi}, \bibinfo{person}{Haoyang Huang},
  \bibinfo{person}{Nan Duan}, \bibinfo{person}{Tianrui Li},
  \bibinfo{person}{Xilin Chen}, {and} \bibinfo{person}{Ming Zhou}.}
  \bibinfo{year}{2020}\natexlab{}.
\newblock \showarticletitle{UniViLM: A Unified Video and Language Pre-Training
  Model for Multimodal Understanding and Generation}.
\newblock \bibinfo{journal}{\emph{ArXiv}}  \bibinfo{volume}{abs/2002.06353}
  (\bibinfo{year}{2020}).
\newblock


\bibitem[Luo et~al\mbox{.}(2022)]%
        {clip4clip2021}
\bibfield{author}{\bibinfo{person}{Huaishao Luo}, \bibinfo{person}{Lei Ji},
  \bibinfo{person}{Ming Zhong}, \bibinfo{person}{Yang Chen},
  \bibinfo{person}{Wen Lei}, \bibinfo{person}{Nan Duan}, {and}
  \bibinfo{person}{Tianrui Li}.} \bibinfo{year}{2022}\natexlab{}.
\newblock \showarticletitle{CLIP4Clip: An Empirical Study of CLIP for End to
  End Video Clip Retrieval and Captioning}.
\newblock \bibinfo{journal}{\emph{Neurocomput.}} \bibinfo{volume}{508},
  \bibinfo{number}{C} (\bibinfo{date}{oct} \bibinfo{year}{2022}),
  \bibinfo{pages}{293–304}.
\newblock
\showISSN{0925-2312}


\bibitem[Ma et~al\mbox{.}(2022)]%
        {xclip2022}
\bibfield{author}{\bibinfo{person}{Yiwei Ma}, \bibinfo{person}{Guohai Xu},
  \bibinfo{person}{Xiaoshuai Sun}, \bibinfo{person}{Ming Yan},
  \bibinfo{person}{Ji~Chao Zhang}, {and} \bibinfo{person}{Rongrong Ji}.}
  \bibinfo{year}{2022}\natexlab{}.
\newblock \showarticletitle{X-CLIP: End-to-End Multi-grained Contrastive
  Learning for Video-Text Retrieval}. In \bibinfo{booktitle}{\emph{Proceedings
  of the 30th ACM International Conference on Multimedia (ACM MM)}}.
  \bibinfo{pages}{638–647}.
\newblock


\bibitem[Mokady et~al\mbox{.}(2021)]%
        {CLIPCAP}
\bibfield{author}{\bibinfo{person}{Ron Mokady}, \bibinfo{person}{Amir Hertz},
  {and} \bibinfo{person}{Amit~H. Bermano}.} \bibinfo{year}{2021}\natexlab{}.
\newblock \showarticletitle{ClipCap: CLIP Prefix for Image Captioning}.
\newblock \bibinfo{journal}{\emph{ArXiv}}  \bibinfo{volume}{abs/2111.09734}
  (\bibinfo{year}{2021}).
\newblock


\bibitem[Patashnik et~al\mbox{.}(2021)]%
        {STYLECLIP}
\bibfield{author}{\bibinfo{person}{Or Patashnik}, \bibinfo{person}{Zongze Wu},
  \bibinfo{person}{Eli Shechtman}, \bibinfo{person}{Daniel Cohen-Or}, {and}
  \bibinfo{person}{Dani Lischinski}.} \bibinfo{year}{2021}\natexlab{}.
\newblock \showarticletitle{StyleCLIP: Text-Driven Manipulation of StyleGAN
  Imagery}. In \bibinfo{booktitle}{\emph{2021 IEEE/CVF International Conference
  on Computer Vision (ICCV)}}. \bibinfo{pages}{2065--2074}.
\newblock


\bibitem[Radford et~al\mbox{.}(2021)]%
        {clip2021}
\bibfield{author}{\bibinfo{person}{Alec Radford}, \bibinfo{person}{Jong~Wook
  Kim}, \bibinfo{person}{Chris Hallacy}, \bibinfo{person}{Aditya Ramesh},
  \bibinfo{person}{Gabriel Goh}, \bibinfo{person}{Sandhini Agarwal},
  \bibinfo{person}{Girish Sastry}, \bibinfo{person}{Amanda Askell},
  \bibinfo{person}{Pamela Mishkin}, \bibinfo{person}{Jack Clark},
  \bibinfo{person}{Gretchen Krueger}, {and} \bibinfo{person}{Ilya Sutskever}.}
  \bibinfo{year}{2021}\natexlab{}.
\newblock \showarticletitle{Learning Transferable Visual Models From Natural
  Language Supervision}. In \bibinfo{booktitle}{\emph{International Conference
  on Machine Learning (ICML)}}.
\newblock


\bibitem[Radovanovic et~al\mbox{.}(2010)]%
        {radovanovic2010hubs}
\bibfield{author}{\bibinfo{person}{Milos Radovanovic},
  \bibinfo{person}{Alexandros Nanopoulos}, {and} \bibinfo{person}{Mirjana
  Ivanovic}.} \bibinfo{year}{2010}\natexlab{}.
\newblock \showarticletitle{Hubs in space: Popular nearest neighbors in
  high-dimensional data}.
\newblock \bibinfo{journal}{\emph{Journal of Machine Learning Research}}
  \bibinfo{volume}{11}, \bibinfo{number}{sept} (\bibinfo{year}{2010}),
  \bibinfo{pages}{2487--2531}.
\newblock


\bibitem[Ramesh et~al\mbox{.}(2022)]%
        {DALL_E2}
\bibfield{author}{\bibinfo{person}{Aditya Ramesh}, \bibinfo{person}{Prafulla
  Dhariwal}, \bibinfo{person}{Alex Nichol}, \bibinfo{person}{Casey Chu}, {and}
  \bibinfo{person}{Mark Chen}.} \bibinfo{year}{2022}\natexlab{}.
\newblock \showarticletitle{Hierarchical Text-Conditional Image Generation with
  CLIP Latents}.
\newblock \bibinfo{journal}{\emph{ArXiv}}  \bibinfo{volume}{abs/2204.06125}
  (\bibinfo{year}{2022}).
\newblock


\bibitem[Schroff et~al\mbox{.}(2015)]%
        {facenet2015}
\bibfield{author}{\bibinfo{person}{Florian Schroff}, \bibinfo{person}{Dmitry
  Kalenichenko}, {and} \bibinfo{person}{James Philbin}.}
  \bibinfo{year}{2015}\natexlab{}.
\newblock \showarticletitle{FaceNet: A unified embedding for face recognition
  and clustering}. In \bibinfo{booktitle}{\emph{2015 IEEE Conference on
  Computer Vision and Pattern Recognition (CVPR)}}. \bibinfo{pages}{815--823}.
\newblock


\bibitem[Sennrich et~al\mbox{.}(2016)]%
        {Sennrich2015NeuralMT}
\bibfield{author}{\bibinfo{person}{Rico Sennrich}, \bibinfo{person}{Barry
  Haddow}, {and} \bibinfo{person}{Alexandra Birch}.}
  \bibinfo{year}{2016}\natexlab{}.
\newblock \showarticletitle{Neural Machine Translation of Rare Words with
  Subword Units}. In \bibinfo{booktitle}{\emph{Proceedings of the 54th Annual
  Meeting of the Association for Computational Linguistics (Volume 1: Long
  Papers)}}. \bibinfo{publisher}{Association for Computational Linguistics},
  \bibinfo{address}{Berlin, Germany}, \bibinfo{pages}{1715--1725}.
\newblock


\bibitem[Sun et~al\mbox{.}(2019)]%
        {videobert2019}
\bibfield{author}{\bibinfo{person}{Chen Sun}, \bibinfo{person}{Austin Myers},
  \bibinfo{person}{Carl Vondrick}, \bibinfo{person}{Kevin~P. Murphy}, {and}
  \bibinfo{person}{Cordelia Schmid}.} \bibinfo{year}{2019}\natexlab{}.
\newblock \showarticletitle{VideoBERT: A Joint Model for Video and Language
  Representation Learning}. In \bibinfo{booktitle}{\emph{2019 IEEE/CVF
  International Conference on Computer Vision (ICCV)}}.
  \bibinfo{pages}{7463--7472}.
\newblock


\bibitem[van~den Oord et~al\mbox{.}(2018)]%
        {infonce2018}
\bibfield{author}{\bibinfo{person}{A{\"a}ron van~den Oord},
  \bibinfo{person}{Yazhe Li}, {and} \bibinfo{person}{Oriol Vinyals}.}
  \bibinfo{year}{2018}\natexlab{}.
\newblock \showarticletitle{Representation Learning with Contrastive Predictive
  Coding}.
\newblock \bibinfo{journal}{\emph{CoRR}} (\bibinfo{year}{2018}).
\newblock


\bibitem[Wang et~al\mbox{.}(2021)]%
        {ActionCLIP}
\bibfield{author}{\bibinfo{person}{Mengmeng Wang}, \bibinfo{person}{Jiazheng
  Xing}, {and} \bibinfo{person}{Yong Liu}.} \bibinfo{year}{2021}\natexlab{}.
\newblock \showarticletitle{ActionCLIP: A New Paradigm for Video Action
  Recognition}.
\newblock \bibinfo{journal}{\emph{ArXiv}}  \bibinfo{volume}{abs/2109.08472}
  (\bibinfo{year}{2021}).
\newblock


\bibitem[Wang et~al\mbox{.}(2022b)]%
        {drl2022}
\bibfield{author}{\bibinfo{person}{Qiang Wang}, \bibinfo{person}{Yanhao Zhang},
  \bibinfo{person}{Yun Zheng}, \bibinfo{person}{Pan Pan}, {and}
  \bibinfo{person}{Xian-Sheng Hua}.} \bibinfo{year}{2022}\natexlab{b}.
\newblock \showarticletitle{Disentangled Representation Learning for Text-Video
  Retrieval}.
\newblock \bibinfo{journal}{\emph{ArXiv}}  \bibinfo{volume}{abs/2203.07111}
  (\bibinfo{year}{2022}).
\newblock


\bibitem[Wang et~al\mbox{.}(2022a)]%
        {Wang2022LearnTU}
\bibfield{author}{\bibinfo{person}{Ziyue Wang}, \bibinfo{person}{Aozhu Chen},
  \bibinfo{person}{Fan Hu}, {and} \bibinfo{person}{Xirong Li}.}
  \bibinfo{year}{2022}\natexlab{a}.
\newblock \showarticletitle{Learn to Understand Negation in Video Retrieval}.
  In \bibinfo{booktitle}{\emph{Proceedings of the 30th ACM International
  Conference on Multimedia (ACM MM)}}. \bibinfo{pages}{434–443}.
\newblock


\bibitem[Wray et~al\mbox{.}(2019)]%
        {finegrained2019}
\bibfield{author}{\bibinfo{person}{Michael Wray}, \bibinfo{person}{Diane
  Larlus}, \bibinfo{person}{Gabriela Csurka}, {and} \bibinfo{person}{Dima
  Damen}.} \bibinfo{year}{2019}\natexlab{}.
\newblock \showarticletitle{Fine-Grained Action Retrieval Through Multiple
  Parts-of-Speech Embeddings}. In \bibinfo{booktitle}{\emph{2019 IEEE/CVF
  International Conference on Computer Vision (ICCV)}}.
  \bibinfo{pages}{450--459}.
\newblock


\bibitem[Wu et~al\mbox{.}(2021)]%
        {hanet2021}
\bibfield{author}{\bibinfo{person}{Peng Wu}, \bibinfo{person}{Xiangteng He},
  \bibinfo{person}{Mingqian Tang}, \bibinfo{person}{Yiliang Lv}, {and}
  \bibinfo{person}{Jing Liu}.} \bibinfo{year}{2021}\natexlab{}.
\newblock \showarticletitle{HANet: Hierarchical Alignment Networks for
  Video-Text Retrieval}. In \bibinfo{booktitle}{\emph{Proceedings of the 29th
  ACM International Conference on Multimedia (ACM MM)}}.
  \bibinfo{pages}{3518–3527}.
\newblock


\bibitem[Xu et~al\mbox{.}(2016)]%
        {msrvtt2016}
\bibfield{author}{\bibinfo{person}{Jun Xu}, \bibinfo{person}{Tao Mei},
  \bibinfo{person}{Ting Yao}, {and} \bibinfo{person}{Yong Rui}.}
  \bibinfo{year}{2016}\natexlab{}.
\newblock \showarticletitle{MSR-VTT: A Large Video Description Dataset for
  Bridging Video and Language}. In \bibinfo{booktitle}{\emph{2016 IEEE
  Conference on Computer Vision and Pattern Recognition (CVPR)}}.
  \bibinfo{pages}{5288--5296}.
\newblock


\bibitem[Yang et~al\mbox{.}(2021)]%
        {taco2021}
\bibfield{author}{\bibinfo{person}{Jianwei Yang}, \bibinfo{person}{Yonatan
  Bisk}, {and} \bibinfo{person}{Jianfeng Gao}.}
  \bibinfo{year}{2021}\natexlab{}.
\newblock \showarticletitle{TACo: Token-aware Cascade Contrastive Learning for
  Video-Text Alignment}. In \bibinfo{booktitle}{\emph{2021 IEEE/CVF
  International Conference on Computer Vision (ICCV)}}.
  \bibinfo{pages}{11542--11552}.
\newblock


\bibitem[Yuan et~al\mbox{.}(2020)]%
        {triplet4reid2020}
\bibfield{author}{\bibinfo{person}{Ye Yuan}, \bibinfo{person}{Wuyang Chen},
  \bibinfo{person}{Yang Yang}, {and} \bibinfo{person}{Zhangyang Wang}.}
  \bibinfo{year}{2020}\natexlab{}.
\newblock \showarticletitle{In Defense of the Triplet Loss Again: Learning
  Robust Person Re-Identification with Fast Approximated Triplet Loss and Label
  Distillation}. In \bibinfo{booktitle}{\emph{2020 IEEE/CVF Conference on
  Computer Vision and Pattern Recognition Workshops (CVPRW)}}.
  \bibinfo{pages}{1454--1463}.
\newblock


\bibitem[Yuhao et~al\mbox{.}(2022)]%
        {zhang2022ArcCSE}
\bibfield{author}{\bibinfo{person}{Zhang Yuhao}, \bibinfo{person}{Zhu Hongji},
  \bibinfo{person}{Wang Yongliang}, \bibinfo{person}{Xu Nan},
  \bibinfo{person}{Li Xiaobo}, {and} \bibinfo{person}{Zhao Binqiang}.}
  \bibinfo{year}{2022}\natexlab{}.
\newblock \showarticletitle{A Contrastive Framework for Learning Sentence
  Representations from Pairwise and Triple-wise Perspective in Angular Space}.
  In \bibinfo{booktitle}{\emph{Proceedings of the 60th Annual Meeting of the
  Association for Computational Linguistics}}. \bibinfo{pages}{4892--4903}.
\newblock


\bibitem[Zhao et~al\mbox{.}(2022)]%
        {centerclip2022}
\bibfield{author}{\bibinfo{person}{Shuai Zhao}, \bibinfo{person}{Linchao Zhu},
  \bibinfo{person}{Xiaohan Wang}, {and} \bibinfo{person}{Yi Yang}.}
  \bibinfo{year}{2022}\natexlab{}.
\newblock \showarticletitle{CenterCLIP: Token Clustering for Efficient
  Text-Video Retrieval}.
\newblock \bibinfo{journal}{\emph{Proceedings of the 45th International ACM
  SIGIR Conference on Research and Development in Information Retrieval
  (SIGIR)}}.
\newblock


\bibitem[Zhou et~al\mbox{.}(2021)]%
        {COOP}
\bibfield{author}{\bibinfo{person}{Kaiyang Zhou}, \bibinfo{person}{Jingkang
  Yang}, \bibinfo{person}{Chen~Change Loy}, {and} \bibinfo{person}{Ziwei Liu}.}
  \bibinfo{year}{2021}\natexlab{}.
\newblock \showarticletitle{Learning to Prompt for Vision-Language Models}.
\newblock \bibinfo{journal}{\emph{International Journal of Computer Vision}}
  \bibinfo{volume}{130} (\bibinfo{year}{2021}), \bibinfo{pages}{2337 -- 2348}.
\newblock


\bibitem[Zhu and Yang(2020)]%
        {actbert2020}
\bibfield{author}{\bibinfo{person}{Linchao Zhu} {and} \bibinfo{person}{Yi
  Yang}.} \bibinfo{year}{2020}\natexlab{}.
\newblock \showarticletitle{ActBERT: Learning Global-Local Video-Text
  Representations}. In \bibinfo{booktitle}{\emph{2020 IEEE/CVF Conference on
  Computer Vision and Pattern Recognition (CVPR)}}.
  \bibinfo{pages}{8743--8752}.
\newblock


\end{thebibliography}

\appendix

\section{Methods}
\subsection{More Details of Feature Representation}
\textbf{Text Encoder.} We adopt the text encoder of CLIP to generate the textual representation, which is a transformer encoder and typically consists of multi-head self-attention (MHSA) and feed-forward (FFN) networks.
Specifically, there are 12 layers and 8 attention heads in the transformer and the query, key, and value features have a 512-dimensional size. The text tokenizer employed in our experiment is a lower-cased byte pair encoding (BPE)~\cite{Sennrich2015NeuralMT} with a 49,152 vocab size. 
After adding a special token $[BOS]$ and $[EOS]$ at the beginning and end of the textual token sequence, respectively, we feed the token sequence into the text encoder to obtain the sentence-level textual feature $\mathbf{t}_{cls} \in \mathbb{R}^{1 \times D}$ and the word-level textual feature $\mathbf{t}_{tokens} = [t_1, t_2, ..., t_M] \in \mathbb{R}^{M \times D}$, where $M$ is the length of $t_i$ and $D$ is the dimension of features. 
The text representations $\mathbf{t}_{cls}$ and $\mathbf{t}_{tokens}$ are outputs of the $[EOS]$ token and corresponding word tokens from the last layer of the text encoder.  

\textbf{Video Encoder.} 
In this work, the video encoder is a standard vision transformer (ViT) with 12 layers, whose architecture is the same as the transformer in natural language processing. The difference is the additional visual tokenization process that turns frames into discrete token sequences. We first sample the given video $v_i$ into $N$ frames with the sampling rate of 1 frame per second (FPS) and convert each frame into $K$ non-overlapped patches. After adding a token $[CLS]$ at the beginning of each token sequence, we feed the token sequence into the video encoder to obtain the frame-level visual feature $\mathbf{f}_{cls} = [f_1, f_2, ..., f_N] \in \mathbb{R}^{N \times D}$ and the patch-level visual feature $\mathbf{v}_{tokens} = [p_{i, cls}, p_{i,0}, p_{i,1}, ..., p_{i,(K-1)}] \in \mathbb{R}^{P \times D}$, where $P=N \times (K+1)$ is the length of the patch sequence. 
The visual representations $\mathbf{f}_{cls}$ and $\mathbf{v}_{tokens}$ are outputs of the $[CLS]$ token and corresponding patch tokens from the last layer of the video encoder. 
Specifically, we use a ViT-B/32 model~\cite{Dosovitskiy2020AnII} with 12 layers and 8 attention heads following the previous work~\cite{clip4clip2021, ts2net2022, xclip2022}.

\subsection{More Details of the Cross-Modal Token Weight Predictor in TPM-CL}
Here we include some further elaboration on the details of the concatenation process described in Eq.~\ref {eq:text_token_weights} in the TPM-CL module.

As discussed in Sec.~\ref {subsec:TPM-CL}, we use a cross-modal feature interaction module to get the weight of each textual token. For a given text $t_i$ and its word-level textual feature $\mathbf t_{tokens}=[t_1,...,t_M] \in \mathbb{R}^{M \times D}$, a video $v_j$ and its frame-level visual feature $\mathbf f_{cls}=[f_1,...,f_N] \in \mathbb{R}^{N \times D}$ , the detailed explanation of the cross-modal token weight predictor is as follows:
\begin{itemize}
\label{itm:xwp_detaila}
    \item First, we apply a dense layer with a trainable weight matrix $W\in \mathbb{R}^{M \times N}$ over the frame-level visual feature $\mathbf{f}_{cls}$ to align its dimension with the word-level textual feature $\mathbf t_{tokens}$:
    \begin{equation}
    \label{eq: align_fcls}
        \mathbf{\hat{f}_{cls}}=W \cdot \mathbf{f}_{cls}=[\hat f_1,...,\hat f_M] \in \mathbb R^{M \times D},
    \end{equation}
    where $\mathbf{\hat{f}_{cls}} \triangleq MLP(\mathbf{f}_{cls})$ in Eq.10. The dense layer is also a lightweight aggregator for the interaction among frames and its weight matrix $W$ is updated during the training phase. 
    
    \item We then concatenate $\mathbf t_{tokens}$ with $\mathbf {\hat{f}_{cls}}$ to get the concatenated word-leval textual feature:
    \begin{equation}
    \label{eq:align_ttokens}
        \mathbf {\hat{t}_{tokens}}=[\hat t_1,...,\hat t_M] \in \mathbb{R}^{M \times D},
    \end{equation}
    where $\mathbf {\hat {t}_{tokens}} \triangleq [\mathbf t_{tokens};MLP(\mathbf{f}_{cls})]$ is the concatenated feature in Eq.~\ref{eq:text_token_weights} and $\hat t_i=[t_i, \hat f_i]$.
    
    \item Finally, we feed the concatenated feature $\mathbf {\hat {t}_{tokens}}$ to an adaptive module $f_{tw}(\cdot)$ to calculate the weight of each textual token.
\end{itemize}

\section{Experiments}
\subsection{More Results with post-processing operations}

The hubness problem~\cite{radovanovic2010hubs,liu2019strong} has been shown to be particularly prevalent in high-dimensional embedding spaces. Qualitatively, the hubness problem means that a small proportion of samples occur disproportionately frequently among the set of k-nearest neighbours of all embeddings~\cite{querybank2022}, which can harm the model's performance. To mitigate the hubness problem observed among cross-modal embeddings for text-video retrieval, some methods adopted Inverted Softmax (IS) to improve the text-video matching. Among them, Dual Softmax loss (DSL)~\cite{camoe2021} and QueryBank Normalization (QB-Norm)~\cite{querybank2022} are two commonly used and effective post-processing operations(such as DRL~\cite{drl2022} uses QB-Norm, and CAMoE~\cite{camoe2021}, TS2-Net~\cite{ts2net2022} and STAN~\cite{liu2023revisiting} use DSL). They can bring significant advancements in performance. 

As shown in Tab.~\ref{tab:res_dsl}, we add the results of our approach with DSL~\cite{camoe2021} to make a fair comparison with previous methods using post-processing operations, \emph{e.g.}, DSL or QB-Norm. Note that our results with DSL still surpass all other methods with significant improvements and achieve SOTA performance on all four datasets. Overall, the good results on different datasets also demonstrate the effectiveness and generalization of our approach.

\begin{table}[]
    \centering
    \caption{Retrieval results with post-processing. $\ast$ means  DSL~\cite{camoe2021} are utilized during inference. $\dag$ denotes re-training.}%
    \label{tab:res_dsl}
    \renewcommand{\arraystretch}{0.8}
    \setlength\tabcolsep{2pt}
    \begin{tabular}{l| c c c c c c}
     \toprule
     \multicolumn{7}{c}{MSR-VTT-1kA}  \\
        \toprule
         {Method} & {R$@$1$\uparrow$} & {R$@$5$\uparrow$} & {R$@$10$\uparrow$} & {MdR$\downarrow$} & 
         {MeanR$\downarrow$} & {rsum$\uparrow$} \\ \toprule   
                \multicolumn{7}{l}{\textit{CLIP-ViT-B/32}} \\ 
       {QB-Norm~\cite{querybank2022}} &47.2 &73.0 &83.0 &2.0 &- &203.2 \\
       {CAMoE$^\ast$~\cite{camoe2021}} &47.3 &74.2 &84.5 &2.0 &11.9 &206.0 \\
        {TS2-Net$^\dag$$^\ast$\cite{ts2net2022}} &50.5	&76.7	&85.9	&1.0	&11.8	&213.1	\\ 
        {STAN$^\ast$~\cite{liu2023revisiting}} &49.0 &74.8 &83.5 &2.0 &- &207.3 \\
        {Baseline$^\ast$} &49.8 &{76.9} &85.8 &2.0 & \textbf{11.3} &212.5 \\
        {Ours$^\ast$} &\textbf{51.7} &\textbf{77.6}  &\textbf{86.2} 	 &\textbf{1.0} 	 &\textbf{11.4} 	 &\textbf{215.5}  \\ 
        \midrule
        \multicolumn{7}{l}{\textit{CLIP-ViT-B/16}} \\  
       {TS2-Net$^\dag$$^\ast$\cite{ts2net2022}} &52.8	&79.0	&87.4	&1.0	&11.4	&219.2\\
       {{DRL~\cite{drl2022}}} &53.3	&80.3	&87.6	&1.0	&-	&221.2	\\
       {STAN$^\ast$~\cite{liu2023revisiting}} &54.1 &79.5 &87.8 &1.0 &- &221.4\\
       {Baseline$^\ast$} &53.3 &78.8&87.1 &1.0& 11.0&219.2 \\
      {Ours$^\ast$} &\textbf{55.5} &\textbf{79.4}& \textbf{87.1} &\textbf{1.0}& \textbf{10.0}& \textbf{222.0} \\ \midrule
      
      \multicolumn{7}{c}{MSVD}  \\
        \midrule
         {Method} & {R$@$1$\uparrow$} & {R$@$5$\uparrow$} & {R$@$10$\uparrow$} & {MdR$\downarrow$} & 
         {MeanR$\downarrow$} & {rsum$\uparrow$} \\ \midrule            
         \multicolumn{7}{l}{\textit{CLIP-ViT-B/32}} \\ 
         {QB-Norm~\cite{querybank2022}} &47.6 &77.6 &86.1 &2.0 &- &211.3\\
        {TS2-Net$^\dag$$^\ast$~\cite{ts2net2022}} & 46.9	&77.3	&85.4	&2.0	&10.5	&209.6 \\
        {{Baseline$^\ast$}} & 46.4	&76.8	&84.6	&2.0	&11.3	&207.8 \\
       {Ours$^{\ast}$} &\textbf{48.7}	&\textbf{78.4}	&\textbf{86.3}	&\textbf{2.0}	&\textbf{9.8}	&\textbf{213.4}  \\ \midrule
        
         \multicolumn{7}{c}{DiDeMo}  \\
         \midrule
          {Method} & {R$@$1$\uparrow$} & {R$@$5$\uparrow$} & {R$@$10$\uparrow$} & {MdR$\downarrow$} & 
         {MeanR$\downarrow$} & {rsum$\uparrow$} \\ \midrule    
         \multicolumn{7}{l}{\textit{CLIP-ViT-B/32}} \\ 
         {QB-Norm~\cite{querybank2022}} &43.5 &71.4 &80.9 &2.0 &-&195.8\\
         {CAMoE$^\ast$~\cite{camoe2021}} &43.8 &71.4 &79.9 &2.0 &16.3 &195.1\\
         {TS2-Net$^\dag$$^\ast$~\cite{ts2net2022}} & 47.1	&73.9	&82.9	&2.0	&12.6	&203.9 \\
        {STAN$^\ast$~\cite{liu2023revisiting}} &51.3 &75.1 &83.4 &1.0 &- &209.8\\
         {{Baseline$^\ast$}} & 49.1	&76.9	&85.1	&2.0	&10.5	&211.1 \\
        {Ours$^{\ast}$} &\textbf{52.7} &\textbf{79.3} &\textbf{86.6} &\textbf{1.0} &\textbf{10.5} &\textbf{218.6}  \\ \midrule
        
         \multicolumn{7}{c}{ActivityNet}  \\
         \midrule
          {Method} & {R$@$1$\uparrow$} & {R$@$5$\uparrow$} & {R$@$10$\uparrow$} & {MdR$\downarrow$} & 
         {MeanR$\downarrow$} & {rsum$\uparrow$} \\ \midrule    
         \multicolumn{7}{l}{\textit{CLIP-ViT-B/32}} \\ 
         {CAMoE$^\ast$~\cite{camoe2021}} &51.0 &77.7 &- &- &- &-\\
         {TS2-Net$^\dag$$^\ast$~\cite{ts2net2022}} & 48.3 	&78.0 	&86.8 	&2.0 	&7.7 	&213.1  \\
        {{Baseline$^\ast$}} & 48.3	&76.3	&86.5	&2.0	&7.6	&211.1 \\
        {Ours$^{\ast}$} &\textbf{53.4}& \textbf{80.7}& \textbf{89.2}& \textbf{1.0}& \textbf{5.3}& \textbf{223.3} \\ 
         \bottomrule
    \end{tabular}
\end{table}

\subsection{More Qualitative Results with TPM-CL}

\begin{figure}[b] 
  \centering
\includegraphics[width=0.9
\linewidth]{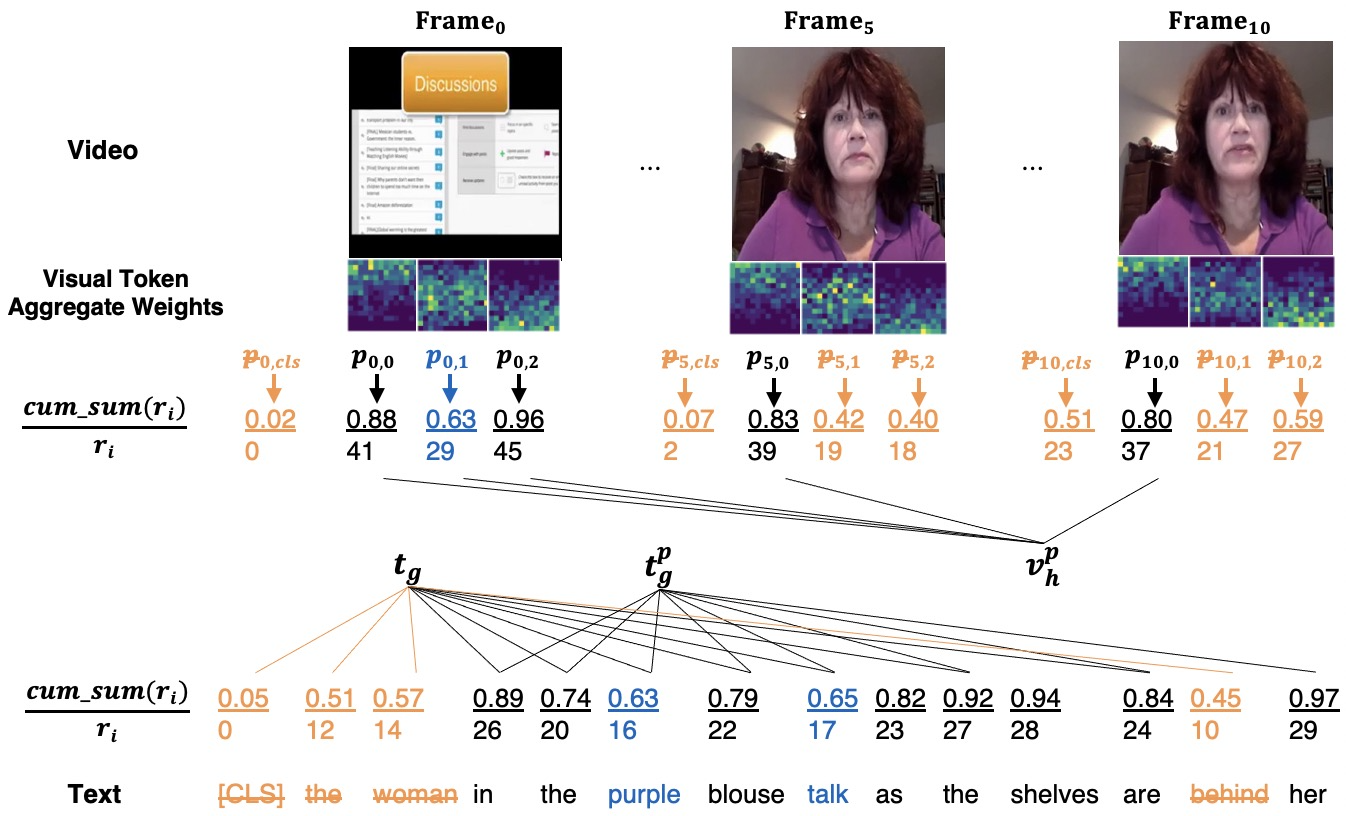}
  \caption{Visualization of the internal mechanism of generating triplet samples in TPM-CL. Masked textual and visual tokens are marked with an orange strikethrough, whose cumulative weights $cum\_sum(r_i)$ are less than $\tau=0.6$. Note that, we give the heatmap visualization of the visual token aggregate weights, with bright yellow colors representing areas of large weights and dark blue colors representing areas of small weights.}
  \label{fig:examples_TPM-CL}
\end{figure}

Fig.~\ref{fig:examples_TPM-CL} shows the internal mechanism of generating the triplet samples. For the given example, the cross-modal interaction masks the informative textual tokens(\emph{i.e.}, \textit{"woman"} and the \textit{[CLS]} token, in the lower half of Fig.~\ref{fig:examples_TPM-CL}) and visual tokens(\emph{i.e.}, the \textit{[CLS]} token for each frame, and the visual tokens $p_{5,1}$ and $p_{5,2}$ in $Frame_5$, in the upper half of Fig.~\ref{fig:examples_TPM-CL}). From the heatmap of the visual token aggregate weights, we can tell that the masked visual tokens $p_{5,1}$ and $p_{5,2}$ are primarily concentrated on the middle and bottom areas of $Frame_5$, which correspond to the female subject.
This indicates that our model has effectively captured the informative tokens and produced accurate, fine-grained hard negatives to represent the subtle semantic differences.


\end{document}